\documentclass[preprints,article,accept,moreauthors]{Definitions/mdpi}

\usepackage{graphicx}

\usepackage{comment}

\firstpage{1} 
\pubvolume{1}
\issuenum{1}
\articlenumber{0}
\pubyear{2026}
\copyrightyear{2026}
\datereceived{ } 
\daterevised{ } % Comment out if no revised date
\dateaccepted{ } 
\datepublished{ } 

\Title{Evaluation Protocols and Validation for Cameras in Indoor Healthcare Monitoring}

\Author{Amirhossein Dadashzadeh $^{1,\dagger}$, Jingjing Liu\textsuperscript{*} $^{2,\dagger}$, Qianhui Men$^{1}$, Qiushuo Cheng$^{2}$, Kirsty Scott$^{3}$, Lisa Alcock$^{3,4}$, Ian Craddock$^{1}$ and Majid Mirmehdi $^{2}$}

\AuthorNames{Amirhossein Dadashzadeh, Jingjing Liu, Qianhui Men, Qiushuo Cheng, Kirsty Scott, Lisa Alcock, Ian Craddock and Majid Mirmehdi}

\address{%
$^{1}$ \quad School of Engineering Mathematics and Technology, University of Bristol, UK  \\
$^{2}$ \quad School of Computer Science, University of Bristol, UK\\
$^{3}$ \quad Translational and Clinical Research Institute, Faculty of Medical Sciences, Newcastle University, UK\\
$^{4}$ \quad NIHR Newcastle Biomedical Research Centre, Newcastle University and The Newcastle upon Tyne Hospitals NHS Foundation Trust, UK\\
}

\corres{Correspondence: jingjing.liu@bristol.ac.uk}

\firstnote{These authors contributed equally to this work and their orders were determined randomly.}
\abstract{Camera-based monitoring systems are increasingly adopted in healthcare settings for the continuous assessment of patient movement and activities. However, their technical performance under real-world indoor conditions remains insufficiently characterised, {preventing appropriate selection when choosing cameras for clinical or home}  adoption and reproducibility. Existing validation studies typically assess either device metrological performance or algorithm accuracy in isolation, and often do not systematically account for practical deployment factors, such as lighting variability, occlusions, and camera positioning. {We present two technical validation protocols: the first evaluates the metrological performance of RGB and RGBD cameras, and the second assesses their use in supporting human pose estimation, validated using state-of-the-art pose estimators}. The proposed protocols systematically assess five cameras (four RGBD and one RGB) under controlled variations in lighting, camera height, viewing angle, and occlusion level, within representative indoor scenarios. The experimental results show that metrological performance varies substantially across cameras, with depth bias at 5m ranging from 50mm to over 1400mm depending on the device. For 2D pose estimation, all cameras achieve broadly comparable accuracy {(mean mAP between \textasciitilde 78\% and \textasciitilde 90\%}) across cameras and estimators, whereas 3D reconstruction error differs markedly across devices (MPJPE ranging from 104mm to 365mm), closely reflecting underlying depth sensing quality. Environmental factors have a camera- and estimator-dependent effect on 3D performance, while camera mounting height has minimal influence within the evaluated range. This work provides evidence-based guidance for the selection and deployment of cameras in healthcare monitoring applications, addressing an important gap in current technical validation practice.}


\begin{document}

%%%%%%%%%%%%%%%%%%%%%%%%%%%%%%%%%%%%%%%%%%
%\setcounter{section}{-1} %% Remove this when starting to work on the template.

\section{Introduction}
The ability to move safely and independently within indoor environments is a key determinant of health, autonomy, and quality of life, particularly for ageing populations and individuals with chronic conditions such as Parkinson’s disease. Changes in mobility and everyday movement patterns are closely linked to disease progression and functional decline, yet are often difficult to capture during short, supervised clinical assessments \cite{studenski2011gait, lord2013moving, masullo2019sit, morgan2020protocol}. Continuous in-home monitoring therefore offers the potential to provide more ecologically valid insights into real-world functional ability \cite{mainsphere, masullo2019sit, morgan2020protocol, jovan2023multimodal,Cheng2025}.

Existing approaches to movement assessment rely primarily on episodic clinical tests or wearable sensing technologies, such as inertial measurement units (IMUs), which have enabled quantitative evaluation of digital mobility outcomes in both laboratory and real-world settings \cite{megaritis2025real, gu2026advancements, ahmed2025advancing}. While wearable sensors provide valuable local kinematic information, they are inherently limited in their ability to capture global spatial context, interactions with the environment, and complex indoor behaviours such as obstacle avoidance, turning around furniture, or sit-to-stand transitions performed in natural settings. {Although IMUs can capture certain activities such as turning and sit-to-stand transitions, their ability to represent full-scene spatial context and interactions remains limited.} In addition, long-term compliance and wearability can pose practical challenges for sustained monitoring in the home \cite{mazza2021technical}.

Camera-based monitoring systems provide a complementary sensing modality for indoor healthcare applications. By enabling markerless, passive observation, cameras can capture rich spatial and contextual information about human behaviour {within indoor environments} without requiring body-worn devices. Recent advances in computer vision, together with the widespread availability of low-cost RGB and RGBD cameras, have accelerated their adoption for applications, including gait analysis, symptom monitoring, and activity recognition and assessment, 
in home and clinical environments \cite{colyer2018review, shaikh2021rgb, heidarivincheh2021multimodal, morgan2023automated, masullo2020person, masullo2021no}. 

%{With increasingly more effort now concentrated on deploying such systems in home environments \cite{morgan2023automated, masullo2021no}, rigorous technical validation of camera performance under realistic indoor conditions has become more critical than ever.}

Despite this growing interest in at-home monitoring \cite{masullo2021no,morgan2023automated}, the technical performance of such cameras under realistic indoor conditions remains insufficiently characterised, leading to poorly informed camera selection. % for clinical and home-based applications. 
{Existing validation studies typically focus on one of two aspects: either to evaluate the metrological performance of camera devices, assessing properties such as depth accuracy, precision, stability, and field of view under controlled conditions \cite{heinemann2022metrological, zennaro2015performance, servi2021metrological} or to evaluate the} accuracy of downstream vision algorithms—most commonly human pose estimation—by comparing derived kinematic or spatiotemporal measures against reference systems such as optical motion capture \cite{lonini2022video, baldinger2025influence, xing2023design, vilas2019validation, d2021validation}. Although a few studies have attempted to combine both perspectives \cite{pfister2014comparative, scataglini2024accuracy}, they typically do not systematically account for practical deployment factors that strongly influence real-world performance, including lighting variability, camera placement, viewing angle, and occlusions. As a result, there is a lack of structured and reproducible technical validation protocols to support the reliable selection and deployment of consumer-grade cameras for indoor healthcare monitoring.

\begin{figure}[t]
\centering
\includegraphics[width=0.9\columnwidth]{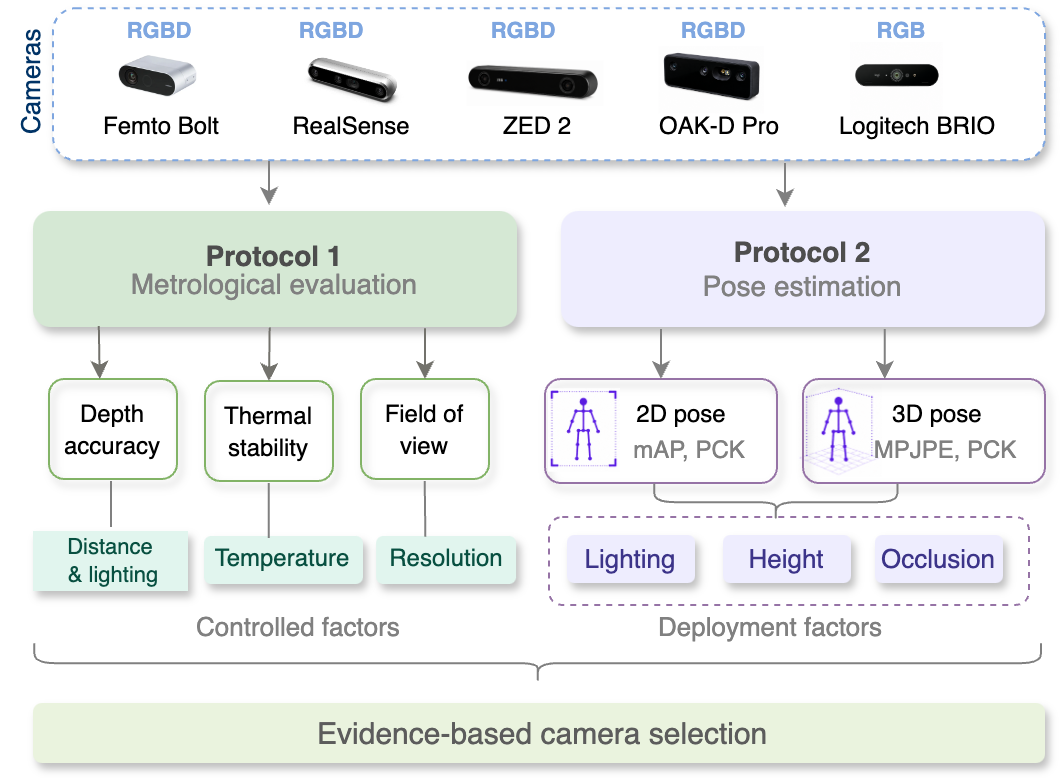}
\caption{Overview of the proposed two-stage validation framework. Protocol~1 evaluates metrological performance across five cameras; Protocol~2 assesses pose estimation accuracy for the cameras, validated against a motion capture system under controlled, indoor deployment conditions. }
\label{fig:roadmap}
\end{figure}

In this paper, we present comprehensive technical validation protocols for cameras that provide evidence-based guidance for their selection and deployment, in particular in indoor-based healthcare monitoring applications {(see Fig. \ref{fig:roadmap})}. The protocols address two complementary aspects of technical validity by assessing sensor-level and application-level performance across realistic deployment conditions: (i) the metrological performance of RGB and RGBD cameras, including depth accuracy, field of view, thermal behaviour, and temporal stability, and (ii) the reliability of these cameras when integrated with human pose estimation under realistic indoor conditions. Five cameras (four RGBD and one RGB) are systematically evaluated in the metrological assessment under controlled variations in lighting, camera placement (height and viewing angle), and occlusion level within representative indoor scenarios.  {For the pose-estimation stage, we considered several candidate estimators and focus on two architecturally distinct, real-time methods, RTMO \cite{lu2024rtmo} and YOLO26 \cite{sapkota2025yolo26}, whose agreement provides a check that camera-level conclusions are not specific to a single algorithm.}

\noindent The main contributions of this work are as follows: 
\begin{itemize}
% \item A two-stage technical validation protocol that integrates camera metrological performance assessment and pose-estimation-based sensing \mmn{I think this contribution can stop here and use the rest of the sentence as a validation contribution, but to be worded such that the validation is done with the given specific outcome.}, including validation against a gold-standard motion capture system \ame{and across  state -of-the-art pose estimators to verify that camera-level conclusions are not algorithm-specific}.
\item {Technical validation protocols for  camera metrological performance assessment and pose-estimation-based sensing for indoor healthcare monitoring.}

\item Validation of these protocols against a gold-standard motion capture system and  architecturally distinct pose estimators.
%, showing that camera-level differences are not estimator-specific.} \mmn{maybe the latter part can be dropped as it's rather obvious?}
\item {A systematically controlled experimental framework that quantifies the impact of key real-world deployment factors, including lighting conditions, camera placement (height and viewing angle), and occlusions.}

\item {Comprehensive empirical analysis and evidence-based insights into the trade-offs between camera accuracy, robustness, and deployment practicality, supporting informed selection of sensing systems for indoor healthcare monitoring.}
\end{itemize}

%%%%%%%%%%%%%%%%%%%%%%%%%%%%%%%%%%%%%%%%%%
\section{Related work}

Camera-based sensing has become an increasingly popular modality for indoor healthcare monitoring, driven by advances in computer vision and the availability of low-cost RGB and RGBD cameras \cite{yeung2019computer,kim2019vision,wang2023contactless,huang2024generalized,zhao2025smarthome,khan2025empowering}. 
%\mmn{4 or 5 citations on indoor healthcare using RGB/D cameras, either from those that appear below in this section or other ones}. 
Existing studies span a wide range of applications, as well as efforts to validate the reliability of camera-based measurements \cite{servi2021metrological,heinemann2022metrological,horsak2024inter, baldinger2025influence}. However, prior work is often fragmented, focusing either on application-level feasibility or on isolated aspects of camera performance. In this section, we review related literature from two complementary perspectives: camera-based indoor healthcare applications, and existing methodologies for validating camera performance.
%\amn{needs a short intro}
\subsection{Camera-based indoor healthcare applications}

With the growing availability of consumer-grade cameras, many studies have explored using RGB and RGBD sensors as low-cost alternatives to marker-based motion capture systems \cite{colyer2018review,lam2023systematic,d2021validation}. 
%Since these cameras can capture movement in \jje{unconstrained environments} \mmn{surely they can capture movement whatever the setting? Also, what would be an unnatural setting?}, 
Their use in healthcare research has reached not just clinical \cite{yeung2019computer,wang2023contactless,huang2024generalized}, but even home environments \cite{kim2019vision,morgan2023multimodal,zhao2025smarthome,khan2025empowering,Cheng2025}. Among these applications, a variety of RGB cameras have been deployed, from  webcams \cite{giuliano2021impersonal,wang2023webcam}, GoPro cameras \cite{gopro1}, and cameras embedded in smartphones or mobile tablets \cite{williams2020supervised, lin2020bradykinesia} to systems such as the Microsoft Kinect \cite{lam2023systematic}, the Intel RealSense \cite{espitia2024development} and the ZED Stereo Camera \cite{valenzuela2025validation}.
%\mmn{be careful, we might be asked to add a Kinect to our work!!}

In healthcare applications, camera sensors are typically used to capture people performing symptom-related motor tasks, including standardized assessments like the Timed Up and Go (TUG) test and {gait trials}. For instance, Dubois et al. \cite{dubois2017automating} use a Kinect to automatically segment the TUG test into sit-to-stand, walking, turning, and stand-to-sit phases. \
%Not sure the following is even relevant here to the discussion
%jje{Average TUG test durations provided by the system differed by only 0.001 s from manual watch-based measurements by clinicians.} \mmn{Is this really correct!?}. \jj{I double checked. It is 0.001s difference while the result is averaged for average difference so some positive and negative values maybe offset.}
Some studies track straight-line walking to estimate spatiotemporal and kinematic metrics such as step length, step width, gait speed, and joint range of motion for evaluating balance and mobility \cite{stenum2021two,cerfoglio2022kinect,cerfoglio2024estimation}. These metrics are usually derived by first recovering body skeletons from RGB or RGBD video using computer vision pose estimation (e.g., OpenPose \cite{cao2019openpose} or Kinect skeletal tracking) and then computing clinically relevant gait and movement indicators.

While these studies demonstrate the potential of camera-based sensing for healthcare monitoring, several limitations remain. First, many existing works primarily focus on application-level feasibility, such as estimating gait parameters or segmenting functional tests, while the reliability of the underlying camera measurements is often assumed rather than systematically validated. Second, although some studies deploy cameras in real-world environments, the influence of environmental variability such as changes in lighting conditions and camera placement has rarely been systematically investigated. Third, different studies employ heterogeneous hardware platforms and pose estimation pipelines, making it difficult to directly compare measurement accuracy across systems. These limitations highlight the need for systematic methodologies to evaluate the performance of camera systems used for indoor healthcare monitoring.

%\amn{This section is very short in compared with B. By the way, is there any limitations of above works we can mention here? }\jj{The next section is the focus of our paper so has more pages.}
\subsection{Existing camera validation methodologies}
%Despite these findings, there is a paucity of research evaluating and validating the performance of such cameras. 
%\sout{Although camera-based sensing has shown promising results in healthcare applications, systematic evaluation of the performance and reliability of these cameras remains relatively limited.}\mmn{Repetition}

{Existing camera validation studies can be broadly categorized into two research directions.} The first focuses on the metrological performance of a device, i.e., the evaluation of its measurement properties such as accuracy, precision, repeatability, stability and geometric fidelity under controlled conditions \cite{carfagni2019metrological,servi2021metrological,heinemann2022metrological, pasinetti2023experimental,abdelsalam2024depth}. This scope of research aims to determine how well a camera, as a sensing instrument, can capture reliable information independent of downstream processing. The second direction examines the validity of features extracted from camera-derived data, where raw RGB or RGBD measurements are combined with pose estimation or motion analysis algorithms, and the resulting kinematic or spatiotemporal metrics are compared against gold-standard references \cite{albert2020evaluation,d2021validation, mehdizadeh2021concurrent,horsak2024inter}. These studies evaluate the overall validity of the sensing and algorithm pipeline in practical application scenarios. Next, we consider existing work in each of these two areas.

For RGBD cameras, an essential component of metrological performance evaluation lies in validating the depth measurement quality. Prior studies \cite{heinemann2022metrological,zennaro2015performance,servi2021metrological} commonly assess depth sensing through a set of fundamental metrological indicators, including bias, precision, and lighting dependence, as well as additional factors such as angle dependent reflectivity and edge precision. These metrics quantify how accurately and reliably a camera can measure depth under varying geometric and environmental conditions. For RGB cameras, prior works \cite{wueller2013low,peltoketo2015mobile,linhares2020good,livada2021low} examined their metrological properties, including image quality, camera speed, lens distortion and low light performance. %{However, RGB imaging is a long-established and mature technology, and standardized evaluation protocols for image quality and camera performance have been widely developed. As RGB imaging primarily captures appearance information rather than direct geometric measurements, in computer vision applications, it is often regarded as a stable visual input rather than a geometric measurement device that requires focused verification. Consequently, most camera validation studies in healthcare sensing focus on the depth sensing performance of RGBD cameras.} 
{However, RGB imaging is a long-established and mature technology with well-developed evaluation protocols, and is often treated as a reliable visual input in computer vision systems. Nevertheless, it is inherently limited to 2D appearance and lacks explicit depth, leading to scale ambiguity in recovering real-world geometry. In healthcare applications such as human pose estimation and gait analysis, accurate geometric measurements (e.g., joint positions, distances, and motion amplitudes) are essential. RGBD cameras address this limitation by providing depth information for recovering absolute 3D geometry. Therefore, depth sensing becomes critical for quantitative evaluation, motivating systematic validation of depth measurement performance.}
%\mmn{This is a nice pgh, but there are two issues. First, how do we know that appearance info is not sufficient for healthcare sensing, and geometric measurements are? What are geometric measurements in this context? What about motion measurements? Secondly, if we take all the reasons above why RGB cameras are not sufficient for healthcare, and RGBD is necessary, it's not clear why RGBD is necessary. I mean the above says: "most validation studies focus on RGBD because RGB is not enough as stated", but it doesn't say why the role of D is important then for evaluation in healthcare applications.}
%\amn{didn't get the last sentence. You mean RGB camera performance is similar because technology is mature?} 

Beyond metrological evaluation of cameras, a second line of research \cite{pfister2014comparative,scataglini2024accuracy} focuses on the concurrent validation of camera-derived data and the algorithms applied to them. These studies typically rely on an additional reference measurement system to establish ground truth. Early work \cite{moreno2017experimental,trinidad2020validation} adopted a single wearable IMU as the reference, which enabled only partial assessment of small range segment motions, such as limited upper-body orientation or trunk/pelvis angular parameters. As validation methodologies matured, researchers increasingly employed optical motion-capture systems such as Vicon \cite{albert2020evaluation}, Qualisys \cite{do2019full}, or OptiTrack \cite{dubois2018validation}, which can capture full-body 3D trajectories at high temporal and spatial resolution and thus serve as the gold standard for evaluating vision based measurements.

Using these reference systems, studies have validated various forms of camera-derived human motion representations. Some works \cite{lonini2022video,baldinger2025influence} extract 2D keypoints directly from a single RGB camera using algorithms such as OpenPose \cite{cao2019openpose} or DeepLabCut \cite{mathis2018deeplabcut}. Others reconstruct 3D joint positions either through multi-camera triangulation of 2D poses \cite{d2021validation} or by leveraging the depth information from RGBD sensors \cite{xing2023design,vilas2019validation}. 
Among these works, data are typically collected under controlled and repeatable movement scenarios. Most commonly, participants walk along a straight overground path \cite{vilas2019validation,lonini2022video,ino2023validity} or on a treadmill \cite{d2021validation} at self-selected or fixed speeds, enabling the extraction of spatiotemporal and kinematic gait parameters. Other studies \cite{xing2023design} focus on static posture assessment, capturing images from multiple viewpoints—such as frontal, lateral, and posterior perspective.

Based on these estimated 2D or 3D poses, researchers compute clinically relevant metrics, particularly those related to neurological and musculoskeletal conditions, including Parkinson's disease \cite{liu2019vision,williams2020supervised,lin2020bradykinesia}, polyneuropathy \cite{vilas2019validation}, stroke \cite{lonini2022video}, and low-back pain \cite{trinidad2020validation}. 
Commonly assessed parameters include spatiotemporal gait features (step length, gait speed, cadence, stance/swing times) \cite{albert2020evaluation, mehdizadeh2021concurrent}, joint kinematics (hip/knee/ankle range of motion) \cite{d2021validation,ino2023validity} and balance-related measures \cite{moreno2017experimental}. 
To quantify agreement between camera-based estimates and the reference system, these studies adopt statistical metrics such as the intra-class correlation coefficient (ICC), Pearson or concordance correlation coefficients, root mean square error (RMSE), coefficient of multiple correlation (CMC), and Bland–Altman analysis. 

{Collectively, these studies demonstrate the feasibility of using low-cost RGB and RGBD cameras for clinical motion assessment, but several limitations remain. 
%Some works have examined the influence of camera placement or height; however, the investigated movement tasks are typically restricted to straight-line walking overground or on a treadmill . In real healthcare scenarios, more diverse behaviours occur, such as sitting and turning. 
{While some works have investigated the influence of camera placement or height \cite{kong2019robust,mehdizadeh2021concurrent}, their evaluations are typically related to specific movements of interest, such as straight-line walking or falling.}
%\mmn{last sentence is confusing how does the 2nd part follow the 1st part?} \jj{I want to say when they evaluate the influence of camera placement, they only consider the motions of their interests.} 
Their influence has not been considered in assessing  common activities such as everyday activities encountered in real-world healthcare settings.
%, particularly in indoor home environments, including sit-to-stand, turns, and carrying out chores. 
%Although such movements have been studied in motion analysis, their impact on camera-based measurement accuracy remains underexplored.} 
%\mmn{many works on sitting/standing exist, again the context here is not clear...is this still related to camera height placement?} \jj{I intended to highlight that these activities are rarely considered in the context of camera validation,}
Moreover, the performance of some RGBD cameras is sensitive to lighting conditions, yet the effect of illumination on human pose estimation has received limited attention. Most existing studies evaluate derived gait or joint kinematic parameters rather than validating the underlying pose data against ground-truth measurements, leaving a gap in understanding how accurately cameras capture raw human motion. 
{Furthermore, while camera manufacturers provide technical specifications under idealized conditions, these measurements are often obtained under controlled laboratory conditions and may not reflect performance in real-world healthcare environments, where factors such as lighting, surface properties, and viewing angles can significantly affect sensing quality \cite{servi2021metrological}. Therefore, independent validation under realistic conditions is necessary.} To address these limitations, {we propose {two technical validation protocols that assess multiple cameras and explicitly consider}} key factors that may affect camera-based motion measurements.}
%%%%%%%%%%%%%%%%%%%%%%%%%%%%%%%%%%%%%%%%%%
\section{Methods }

Indoor deployment of cameras (particularly at home for healthcare monitoring) is strongly affected by practical factors such as lighting variability, occlusions, and movement characteristics, which we treat as controlled experimental variables. Next, we propose %{a two-stage evaluation protocol}
 evaluation protocols that (i) quantitatively {characterise} camera metrological performance and (ii) {assess} camera suitability for pose-estimation-based monitoring. {These protocols provide} a reproducible and application-oriented framework for benchmarking camera systems under clinically relevant indoor conditions.
%\subsection{Method overview}
% \subsection{Protocol 1: Verification of  metrological performance}

% This protocol evaluates the metrological performance of camera systems across three complementary aspects: (i) depth measurement error, (ii) thermal behaviour and temporal stability, and (iii) field of view (FOV). 

% \mmn{To avoid repetition, blend this next pgh into the corresponding subsections below or just remove it if it's not totally necessary and the content is already covered elsewhere:} Depth measurement error is assessed to characterise the accuracy and precision {of the depth sensing modality. Following existing works \cite{corti2016metrological}, bias is defined as the mean difference between the measured depth and the ground-truth distance, reflecting systematic error, while precision is defined as the variability of repeated measurements, quantified by the standard deviation of the depth estimates.}

% \mmn{To avoid repetition, blend this next pgh into the numbered, corresponding subsections below or just remove it if it's not totally necessary and the content is already covered elsewhere:} Thermal and stability analysis examines performance variations over time and under thermal changes, which are critical for long-term, unattended indoor deployment. Finally, field-of-view measurements quantify the spatial coverage of each device, directly affecting scene observability and the reliability of subject tracking. {The evaluated devices include four RGBD cameras (Femto Bolt, Intel RealSense D456, ZED 2, and OAK-D Pro) and one RGB camera (Logitech BRIO 4K).} 

\subsection{Protocol 1: {Evaluation of metrological performance}}
{This protocol evaluates the metrological performance of camera systems across three complementary aspects: (i) depth measurement error, (ii) thermal behaviour and temporal stability, and (iii) field of view (FOV). The evaluated devices include four RGBD cameras (Femto Bolt, Intel RealSense D456, ZED 2, and OAK-D Pro) and one RGB camera (Logitech BRIO 4K).}

\begin{figure}[t]
    \centering
    \includegraphics[width=0.8\columnwidth]{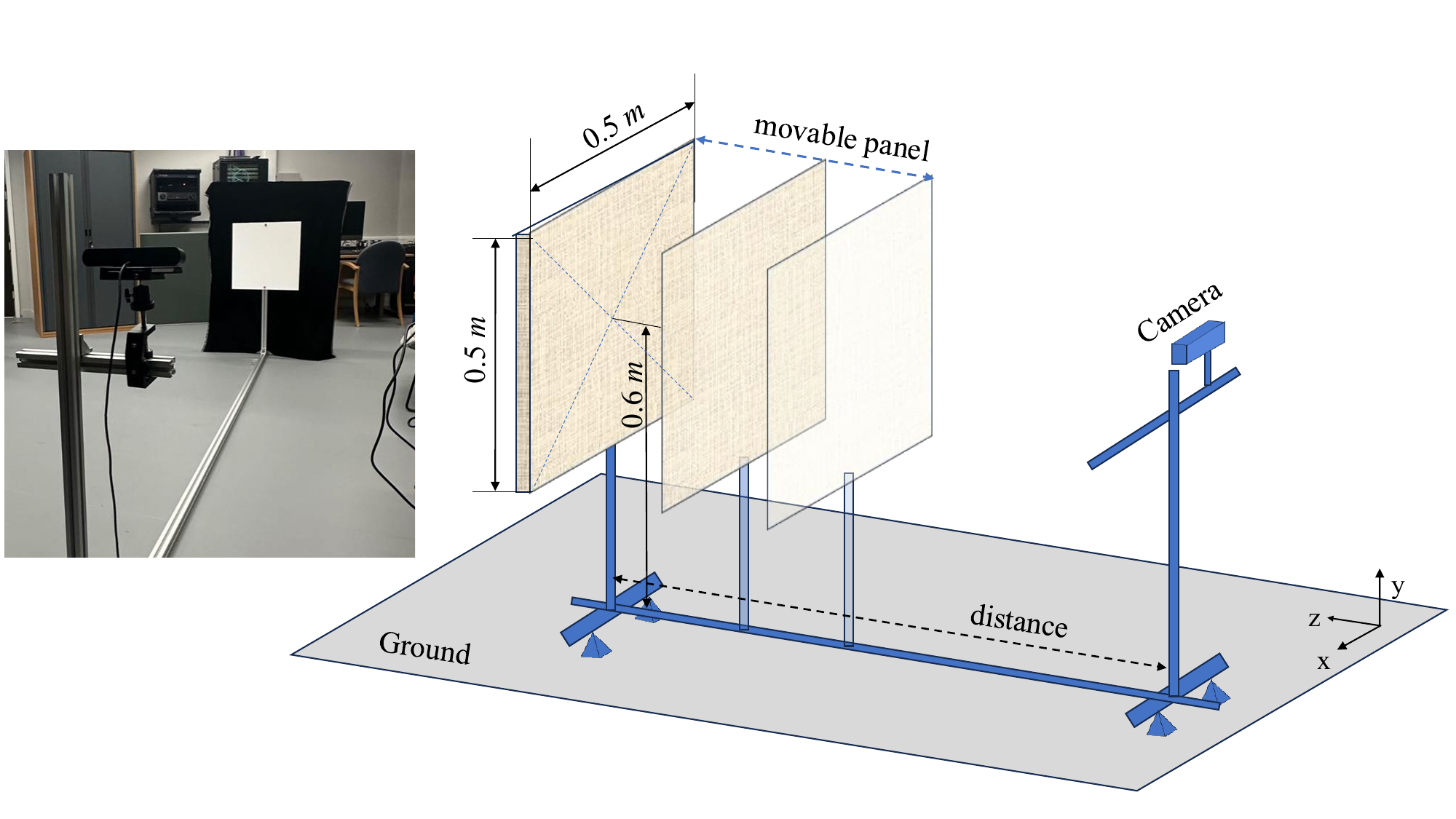}
    \caption{Experimental platform setup for measuring depth error. }
    \label{fig:pf_depth}
\end{figure}

{\bf Depth measurement error -}
Following established metrological evaluation methodologies for RGB-D cameras \cite{heinemann2022metrological, zennaro2015performance}, this protocol assesses depth measurement error by measuring the distance between the camera and a planar target under controlled conditions. The target is positioned perpendicular to the camera’s optical axis $z$ to ensure that measured depth values correspond directly to the camera–plane distance. The cameras and two laser range finders {(RockSeed S2)}, used as ground-truth references, are mounted on construction profiles aligned in the $xy$ plane. A schematic of the experimental platform is shown in Fig.~\ref{fig:pf_depth}.  %\amn{maybe keep it for the appendix? we can keep either figure 1 here or our actual figure.}
This configuration provides a controlled and repeatable setup for quantifying depth sensing accuracy across different camera devices.
Depth measurements are evaluated under controlled variations of both lighting conditions and camera-to-object distance. Ambient illumination is monitored using a lux meter to ensure consistency across experiments. Under normal indoor lighting conditions (450–600 lux), depth is recorded at distances ranging from $1$ to $5m$ in $1m$ increments. {To assess the influence of reduced illumination, depths are measured under two scenarios: a low-light condition (50–100 lux) and a very low-light condition (3–10 lux). For both, measurements are again acquired at $1$ to $5m$ distances.} %, consistent with the normal lighting setup.}

\subsubsection{Thermal behaviour and temporal stability}
Thermal behaviour and temporal stability are critical for camera systems intended for continuous and unattended indoor monitoring, with measurement reliability affected over time due to   temperature-induced sensor drift~\cite{vila2021method}. {To characterise these, we evaluate camera stability by continuously operating each device for two hours while monitoring depth measurements and device temperature, with the target positioned at a fixed distance of $2m$.

% \mmn{I would guess  referees will object that 2 hours is not enough.}

}

Depth drift is quantified by measuring the variation in the mean depth value over time. At each recorded time step $t$ (sampled at 30~fps), the mean depth $\bar{d}(t)$ is computed over a central region of interest consisting of 500~pixels, selected to minimise edge effects and represent the most stable portion of the depth image. Let $\bar{d}(0)$ denote the mean depth at the first recorded time step. Two complementary metrics are used to characterise depth stability over a total of $N$ recorded time steps. We apply RMSE  to calculate the overall deviation of the depth signal from the initial measurement, 
\begin{equation}
    \text{RMSE} = \sqrt{\frac{1}{N} \sum_{t=1}^{N} \left( \bar{d}(t) - \bar{d}(0) \right)^2},
    \label{eq:rmse_stability}
\end{equation}
and the average absolute drift (AAD) to quantify the mean magnitude of frame-to-frame depth variation:
\begin{equation}
    \text{AAD} = \frac{1}{N-1} \sum_{t=1}^{N-1} \left| \bar{d}(t+1) - \bar{d}(t) \right|.
    \label{eq:aad_stability}
\end{equation}
RMSE captures the cumulative deviation from the initial reference, reflecting both systematic drift and random noise, while AAD characterises the short-term frame-to-frame jitter of the depth measurement. Both metrics are recorded in millimetres. As electronic imaging sensors and signal-conditioning circuits are sensitive to temperature changes, these metrics provide indicators of thermally induced measurement instability.

{Device temperature is monitored concurrently throughout each operating session} using an external thermal camera (Sealey VS913). The maximum observed surface temperature of each device is noted as the representative device temperature at each time point.
Three experimental factors are considered,  scene dynamics, {camera operation}, and ambient room temperature. These are examined through two separate experimental configurations as described below.

\textit{{Scene dynamics and camera operation:}} {Cameras are deployed in the natural indoor environment of a home setting, where illumination and scene content may vary due to normal human activity. {Cameras are operated continuously for two hours, during which a thermal camera} records device surface temperature profiles.}

%\mmn{What exactly re scene dynamics are tangibly/quantifiably examined? It's not clear from the above?}

\textit{Ambient temperature conditions:} To isolate the effect of environmental temperature, cameras are mounted on the experimental platform at a fixed distance of $2m$ from the planar target, with controlled lighting and no people in the field of view. An indoor air-conditioning system is used to regulate room temperature across three ranges: 18--20$^\circ$C (low), 23--25$^\circ$C (normal), and 28--30$^\circ$C (high). For each condition, cameras operate at their maximum supported frame rate and resolution, while both device temperature and depth drift are recorded over time.

\subsubsection{Field of view angles}
The FOV of a camera defines the angular extent of the observable scene and directly determines the spatial coverage achievable by the sensing system. A larger FOV enables broader scene coverage, which is particularly important for indoor monitoring scenarios where subject movement may span across large widths in the scene.

\begin{figure}[]
    \centering
\subfloat[\centering]{\includegraphics[width=7.5cm]{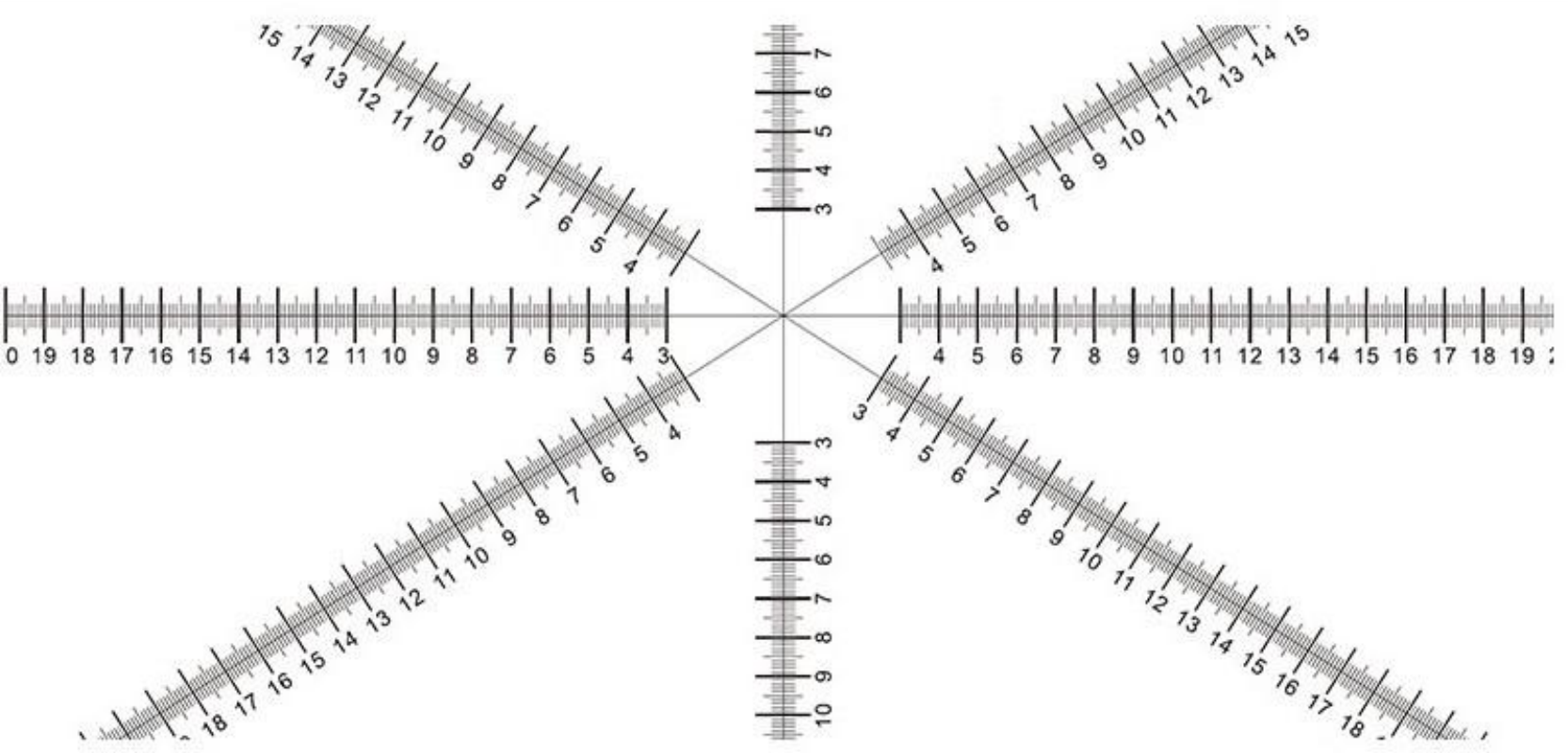}}
\subfloat[\centering]{\includegraphics[width=5cm]{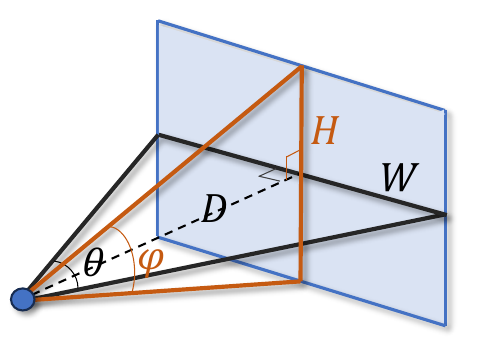}}
    \caption{FOV angle measurements. (a) Acrylic sheet for FOV. (b) FOV angle calculations.} %\mmn{Make the text font in part (b) bigger. Replace L with H, and H with V? The red brackets are unnecessary.}}
\label{fig:cal_fov}
\end{figure}

As illustrated in Fig.~\ref{fig:cal_fov}a, an acrylic sheet with annotated dimensions is mounted on the planar surface of the experimental platform. The camera is carefully positioned to ensure that the image boundaries are aligned with the edges of the acrylic sheet and that the viewing direction is approximately perpendicular to the sheet. Images are acquired at a fixed shooting distance (e.g.\ $50cm$), which is measured using a laser distance metre. The horizontal and vertical extents of the captured area are then determined based on the ruler markings on the sheet.
As shown in Fig.~\ref{fig:cal_fov}b, the horizontal and vertical field-of-view angles are computed from the measured dimensions using
\begin{equation}
    \theta = 2\tan^{-1}\!\left(\frac{W}{2D}\right), \qquad 
    \varphi = 2\tan^{-1}\!\left(\frac{H}{2D}\right),
\end{equation}
where $W$ and $H$ denote the measured width and height of the visible area on the sheet, respectively, and $D$ is the camera-to-plane distance.

\subsection{Protocol 2: Validation of human pose estimation performance}

{Participants are asked to perform various motion tasks while their movements are simultaneously recorded using cameras and a Vicon optical motion capture system {(8 Vantage v5 cameras and 10 Vero v2.2 cameras)}. This will facilitate quantitative comparison of pose estimation results against the mocap system, which serves as the reference measurement standard.

The following subsections describe the instrumentation, experimental setup, data collection, signal processing and synchronisation, spatial alignment of skeleton data, as well as the evaluation metrics used in this study.

\subsubsection{Instrumentation}
Four cameras are evaluated for indoor human pose estimation, including three RGBD cameras, namely Femto Bolt, Intel RealSense D456, and ZED 2, as well as one RGB camera, Logitech BRIO 4K. 
Following the depth evaluation experiment  presented later, the RGBD camera OAK-D Pro was not included in other experiments, as it was found to exhibit substantially larger depth measurement errors compared with the other RGBD cameras.

% \mmn{I think we should say THREE cameras in the Introduction, if you  are really not dealing much with OAKD. Unless you have results for it anyway and just include it everywhere and remove the last declaration. }

\subsubsection{Experimental setup and data collection}
\label{Sec:Exp_data}
Experiments were conducted with a single healthy male participant (age: 32 years; height: 176 cm; body mass: 84 kg). A single participant was used to ensure consistent body geometry and motion patterns across repeated trials, allowing the evaluation to focus on the measurement characteristics of the camera systems rather than inter-subject variability.

The participant was  equipped with a full-body set of 26 retro-reflective markers. To ensure compatibility with the detected joints' positions in the 2D human pose estimation method, e.g., RTMO \cite{lu2024rtmo}, we created our own body template with the specified positions of the markers shown in Fig.~\ref{fig:template}. %\amn{where did we borrow this picture?} \jj{from GPT} 
This template covers anatomically meaningful joint locations, including the head, shoulders, elbows, wrists, hip center, knees, and ankles, {which are commonly used in human motion analysis and have been shown to be particularly relevant for functional movements, such as turning \cite{Cheng2025}.} %\mmn{cite Qiushuo's paper "Your Turn..." where he briefly studied the most significant joints for turning}. 
{3D marker trajectories were captured at 100Hz using the Vicon system consisting of 17 infrared cameras. The data were acquired and processed using Nexus software.} %\amn{repetition of "Vicon". The last few sentence needs to be re-wrriten.}

\begin{figure}[h]
    \centering
        \includegraphics[width=0.98\columnwidth]{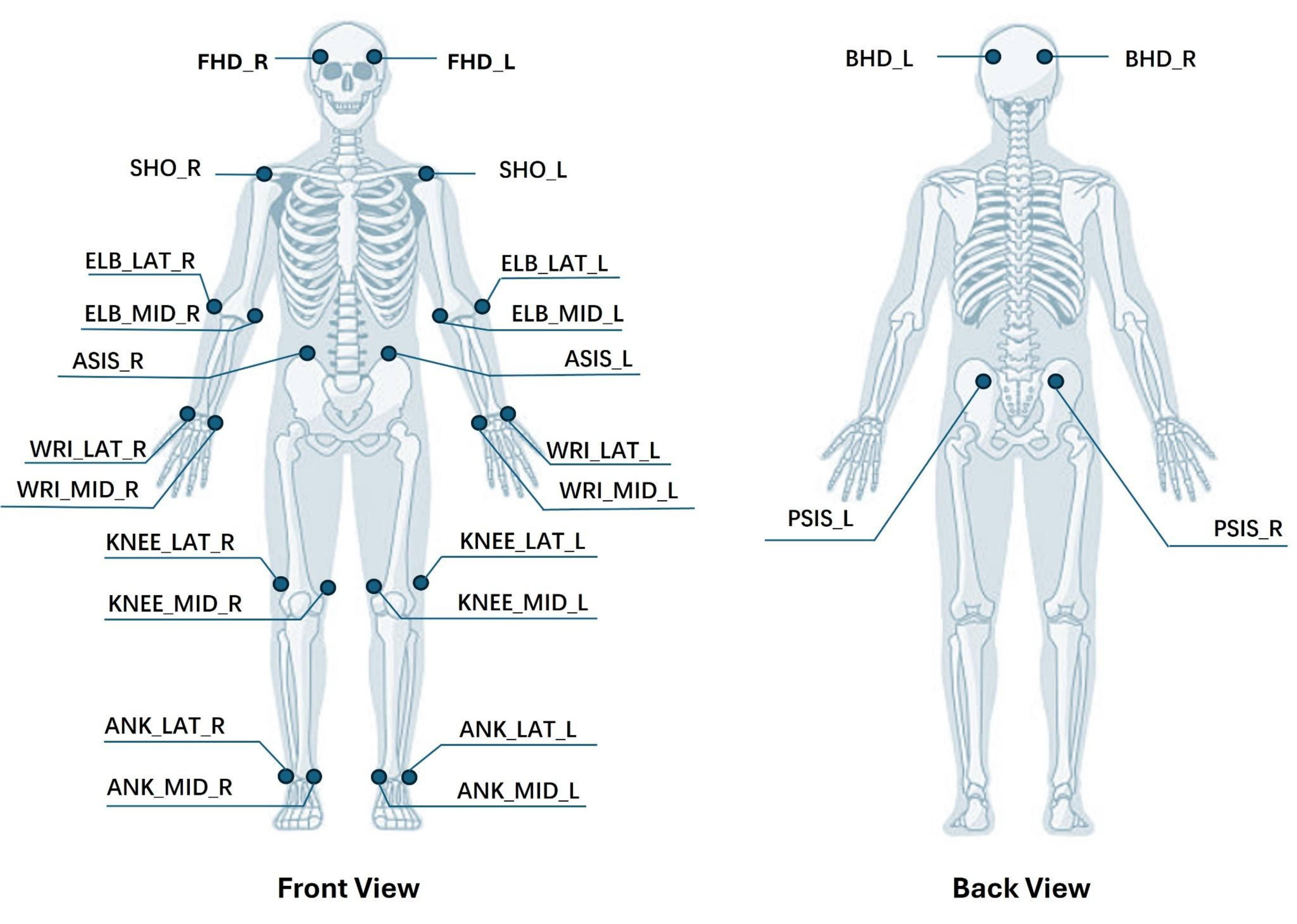}
    \caption{Template of marker arrangement used for the motion capture system, shown in front (left) and back (right) views. Marker labels use the following abbreviations: FHD (forehead), BHD (back of head), SHO (shoulder), ELB (elbow), WRI (wrist), KNEE (knee), and ANK (ankle). The suffixes L and R denote left and right sides of the body, respectively, 
while LAT and MID indicate lateral and medial marker positions. 
This marker configuration is designed to enable consistent alignment between motion capture data and camera-based human pose estimation.} %\mmn{needs more description and definition of the acronym and shortened label names}}
\label{fig:template}
\end{figure}

As illustrated in Fig.~\ref{fig:experimental_setup}, the participant was instructed to follow a predefined path within the experimental area and perform a sequence of functional tasks commonly encountered in daily indoor activities. The task sequence includes sitting, sit-to-stand, walking along straight lines, in-place turning, and continuous walking with directional changes. The participant completed all tasks at a self-selected comfortable pace to reflect natural movement patterns, without external constraints on speed or cadence.

\begin{figure}[htp]
    \centering
    \includegraphics[width=7cm]{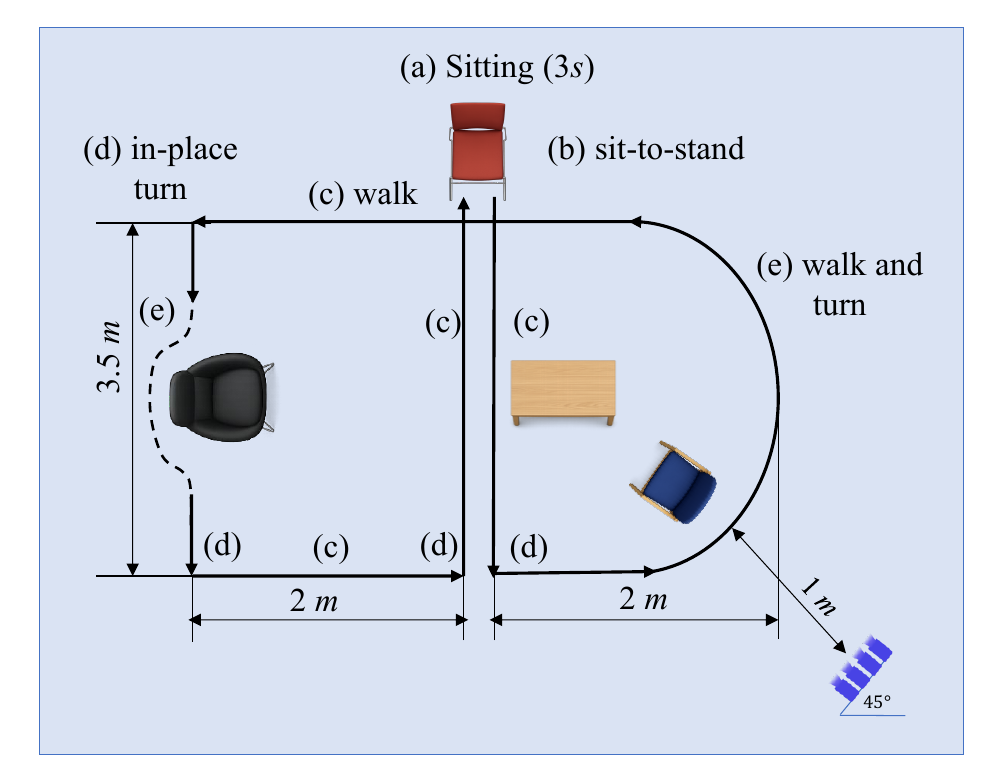}
    \includegraphics[width=5.5cm]{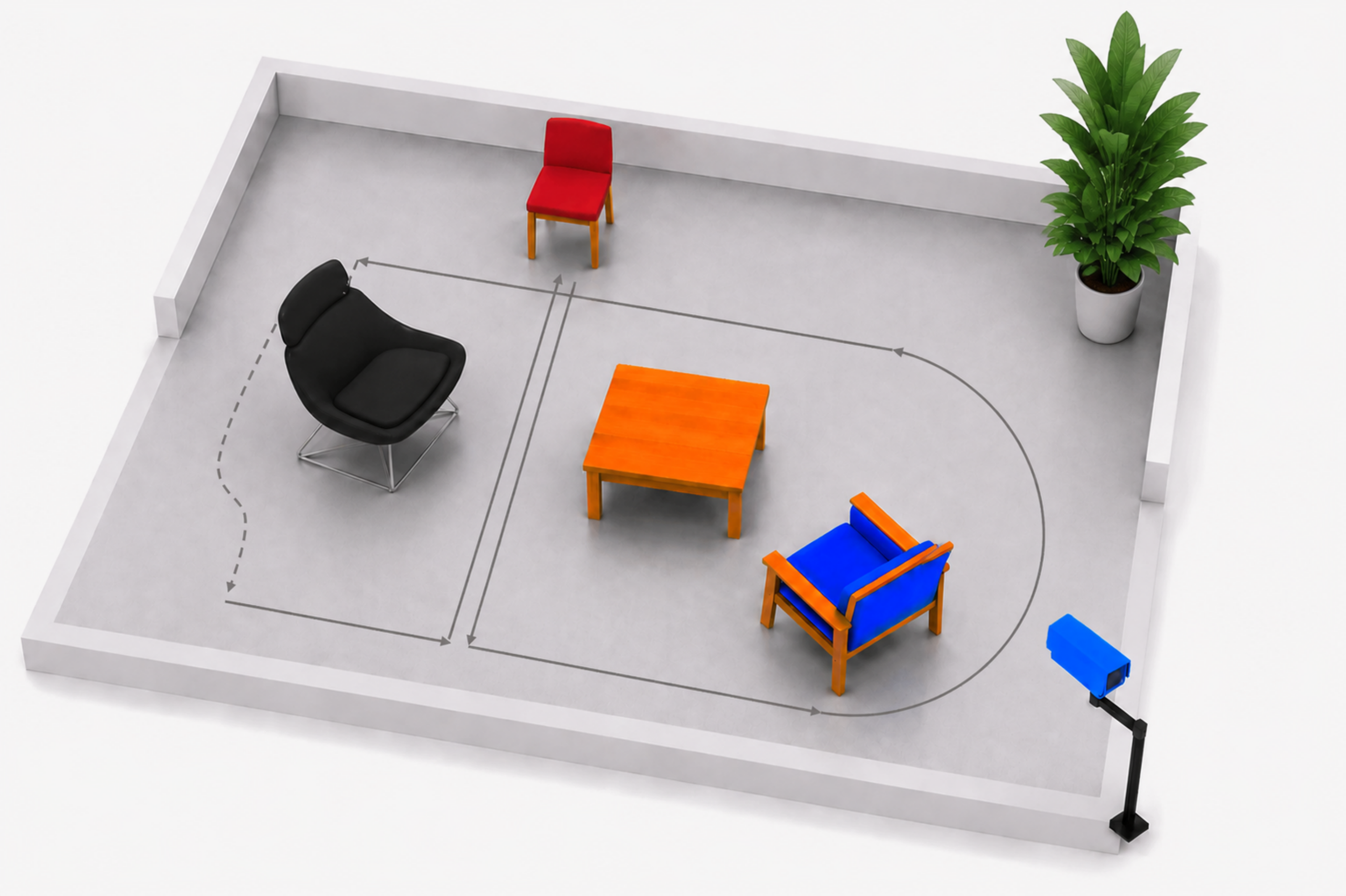}
    \caption{
    Experimental platform setup and motion task used for evaluating human pose estimation performance. The participant started from the chair and performed a sequence of motion tasks including (a) sitting for 3 s, (b) sit-to-stand, (c) straight walking along the designated path, (d) in-place turning, and (e) walking with a curved turn before returning to the starting point. The layout also shows the spatial arrangement of furniture (e.g., chairs and coffee table), the walking paths, and the camera placement relative to the capture area. }
    %\mmn{A sofa is usually for at least 2 people. The image there is an Armchair.} \jj{As shown in Fig.5, can we say it a single-sofa? Or else it sounds similar to the other chairs.} }
    \label{fig:experimental_setup}
\end{figure}

As shown in Fig.~\ref{fig:camera_setup}, the Vicon cameras were distributed around the perimeter of the experimental area to ensure full coverage of the capture volume. The four cameras under investigation were mounted on tripods and oriented toward the capture space.

\begin{figure}[h]
    \centering
    \includegraphics[width=0.7\columnwidth]{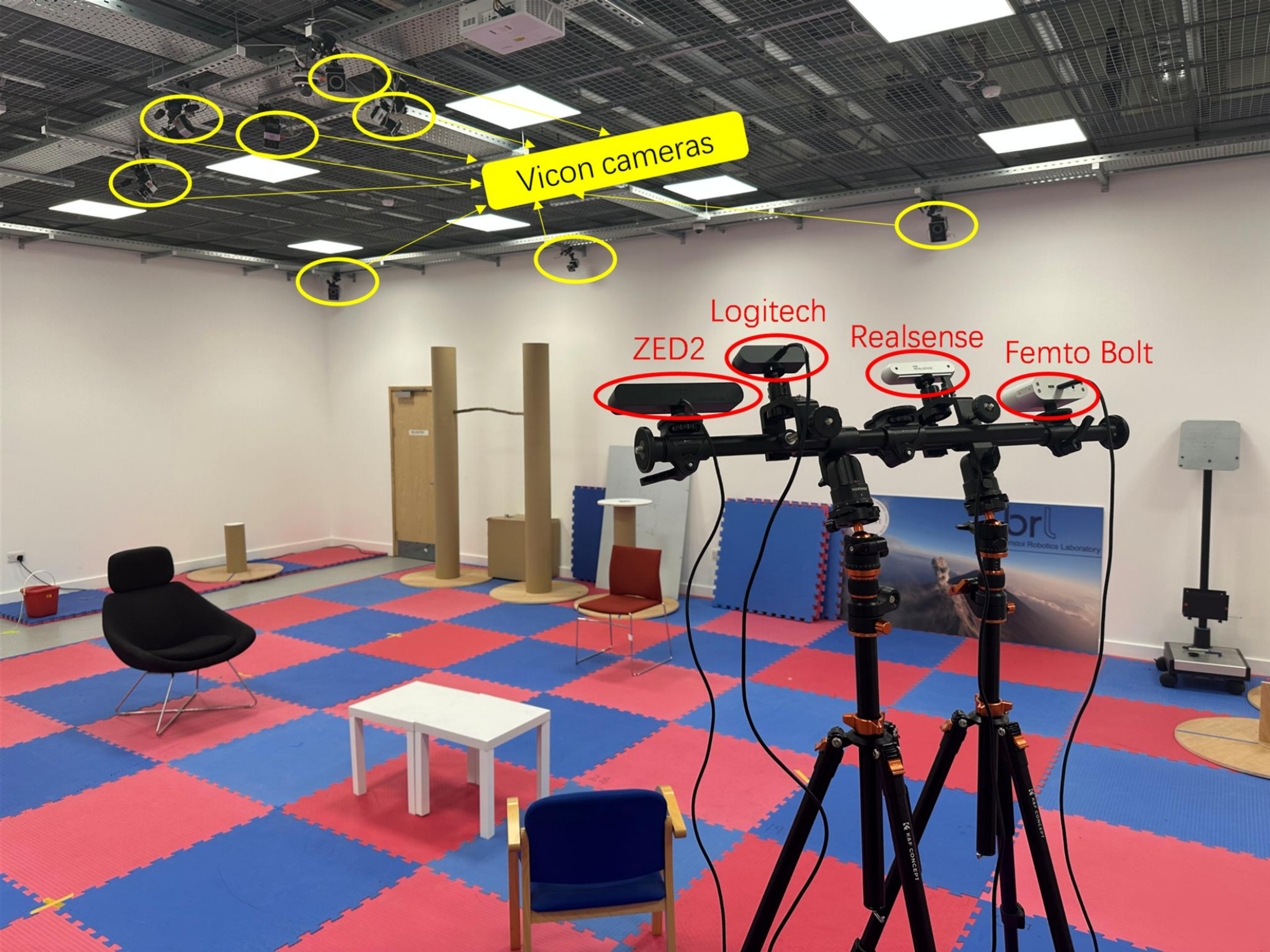}
    \caption{Experimental setup of cameras and Vicon motion capture system.}
    \label{fig:camera_setup}
\end{figure}

Three control factors were considered in this experiment: lighting condition, camera location, and occlusion scale, as summarised in Table \ref{control_factor}. Two lighting conditions were evaluated, normal lighting (450-600lux) and low-lighting (50-100lux). Camera location was examined at three mounting heights %($1.8m$, $2.0m$, and $2.2m$), 
with corresponding viewing angles (explained later) selected for each height. 
Finally, two occlusion conditions were considered: a non-occluded condition without any furniture and an occluded condition with furniture present in the capture area. {The participant repeated the full motion task under every combination of these experimental control factors.} %\amn{we need to clarify whether the participant repeated the full task protocol under every combination of lighting, camera height, and occlusion condition.}

\begin{table}[H]
\centering
\caption{Controlled experimental variables and their corresponding levels.}
\label{tab:control_factors}
\begin{tabular}{ll}
\toprule
\textbf{Control Factor} & \textbf{Levels} \\
\hline
Lighting condition
& Normal lighting \\
& Low lighting \\
\hline
Camera location
& $1.8m$ \\
& $2.0m$ \\
& $2.2m$ \\
\hline
Occlusion
& No \\
& Yes \\
\bottomrule
\end{tabular}
\label{control_factor}
\end{table}

To ensure that the full human body remains within the camera field of view, the camera must be positioned at an appropriate tilt angle. For a given camera height, the tilt angle is determined by the requirement that the field of view fully contains both the subject's head and feet, {corresponding to the limiting cases where the head or feet lie exactly on the field-of-view boundaries.} Based on this geometric relationship, as illustrated in Fig.~\ref{fig:cal_ota}, the allowable tilt angle range is defined by corresponding upper and lower bounds.
{Specifically, the camera tilt angle $\beta$ is restricted to
\begin{equation}
\beta_{lower}-\frac{\varphi}{2} < \beta < \beta_{upper}+\frac{\varphi}{2},
\label{eq:tilt_angle}
\end{equation}
where $\beta_{lower}=\tan^{-1}\left(\frac{h_c}{d}\right)$ and $\beta_{upper}=\tan^{-1}\left(\frac{h_c-h_o}{d}\right)$. Here, $h_c$ denotes the camera mounting height, $h_o$ the subject height, $d$ the horizontal distance between the camera and the subject, and $\varphi$ the vertical field of view of the camera.} The distance $d$ is assumed to satisfy $m \leq d \leq m+n$, {where $m$ and $m+n$ denote the near and far boundaries of the expected observation region, respectively.} 
In the experimental setup, the tilt angle ranges are computed for each camera at three mounting heights ($1.8m$, $2.0m$, and $2.2m$), {which are selected to reflect typical indoor installation scenarios, such as wall-mounted or elevated camera placements in residential and clinical environments} 
%\mmn{probably should explain why these heights are used}. 
The subject height is set as $h_o = 1.8m$, and the camera-to-subject distance is assumed to lie within the interval $1.5m \leq d \leq 5.0m$, covering typical indoor interaction distances. For each camera and mounting height, the corresponding feasible tilt angle range is calculated using Eq. (\ref{eq:tilt_angle}), and the recommended 
%\amn{why midpont is optimal? need to be explained here.}  
tilt angle is defined as the midpoint of this range. {This choice provides a representative tilt angle within the feasible range, providing a balanced margin with respect to both the upper and very low visibility constraints.} 

\begin{figure}[t]
    \centering
    \includegraphics[width=0.65\columnwidth]{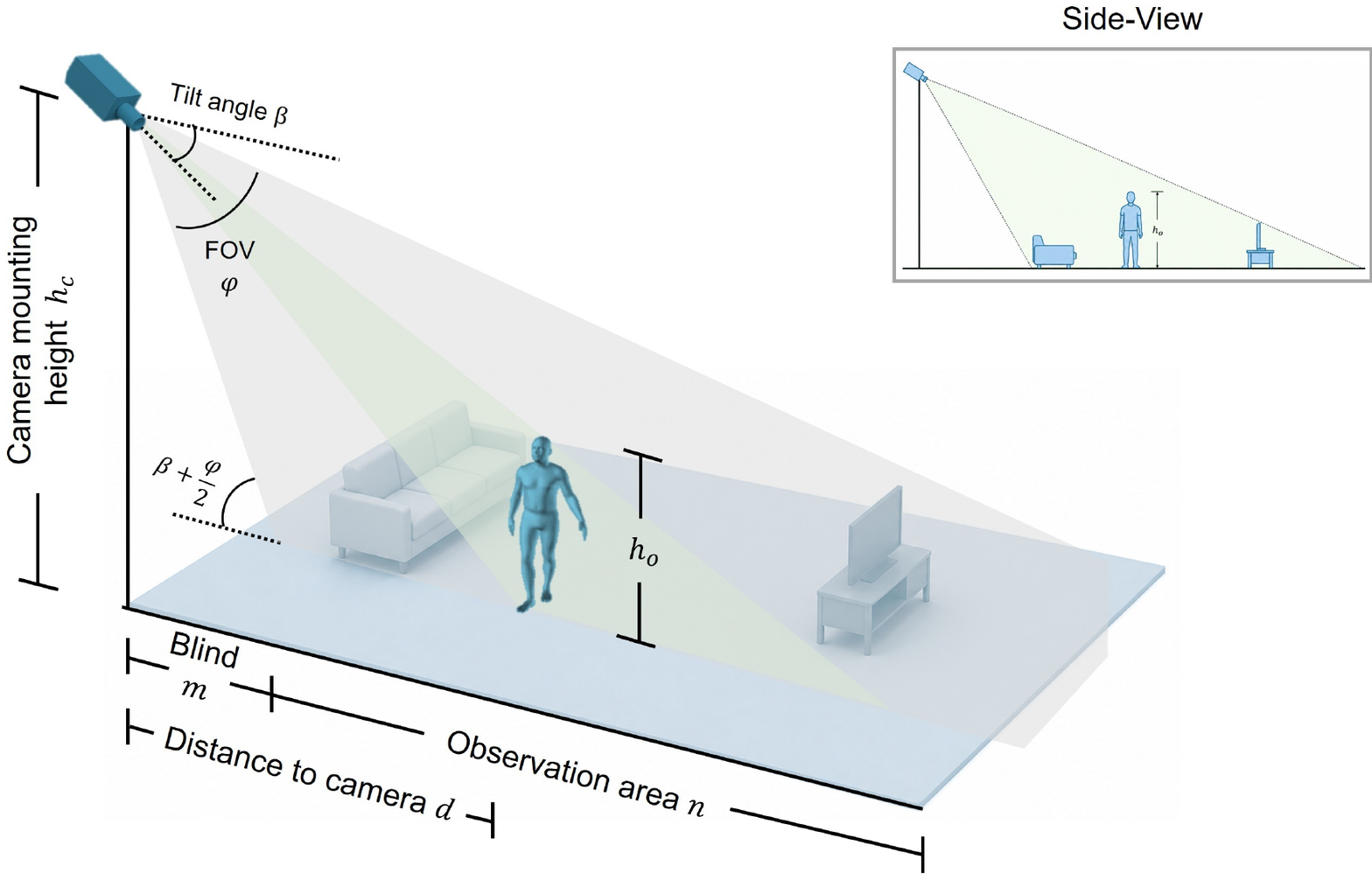}
    %\begin{tikzpicture}
    %    \node[inner sep=0] (img)
    %{\includegraphics[width=0.65\columnwidth]{images/optimal_tilt_angle_3d_smpl_side.pdf}};
    %    \node[black,font=\bfseries\small]
        %at ($(img.north east)+(-1.0cm,-0.5cm)$)
        %{Side-view};
    %\end{tikzpicture}
    \caption{Demonstration of the feasible camera tilt angle range measurement. }
    \label{fig:cal_ota}
\end{figure}

The resulting angle ranges and recommended tilt angles are summarised in Table~\ref{tab:tilt_angle}. For the Femto Bolt at a mounting height of $2.2m$, there is no solution for the computed range due to field-of-view limitations. %\amn{we need to discuss the implication of using a non-feasible tilt angle for Femto Bolt at 2.2m as this may affect fairness of cross-camera comparison.} 
To maintain consistency across cameras, a tilt angle of $30^\circ$ is adopted for this configuration. {Although this angle lies outside the theoretically feasible range, it still keeps most of the subject within the camera field of view.}
% repetition: and highlights the field-of-view limitation of the Femto Bolt at this mounting height.}

\begin{table}[H]
\centering
\caption{The resulting tilt angle (\textdegree) for the four cameras under three mounted heights.}
%\resizebox{1.0\columnwidth}{!}{
\begin{tabular}{c|cc|cc|cc|cc}
\toprule
\textbf{Camera} & \multicolumn{2}{c|}{\textbf{ZED2}} & \multicolumn{2}{c|}{\textbf{RealSense}} & \multicolumn{2}{c|}{\textbf{FemtoBolt}} & \multicolumn{2}{c}{\textbf{Logitech BRIO 4K}} \\ \cline{2-9}
                & Range               & Avg.         & Range                  & Avg.           & Range               & Avg.              & Range                 & Avg.          \\ \hline
1.8m            & {[}16,35{]}         & 26           & {[}18,32{]}            & 25             & 25                   & 25                & {[}25,26{]}           &      25         \\
2.0m            & {[}19,37{]}         & 28           & {[}21,34{]}            & 28             & 27                   & 27                & 28                     & 28            \\
2.2m            & {[}21,39{]}         & 30           & {[}24,37{]}            & 30             & -               & 30                & 30                     & 30            \\ \bottomrule
\end{tabular}
%}
\label{tab:tilt_angle}
\end{table}
\subsubsection{Signal processing and synchronisation}
%{Data post-processing includes marker labeling and reconstruction of the full-body motion model. Marker trajectories are initially labeled automatically and subsequently inspected to ensure consistency. Temporal gaps in the marker data caused by occlusions are manually filled within the Vicon Nexus software to for subsequent analysis.}
Data post-processing includes labelling marker trajectories automatically to start with and then inspecting them to ensure consistency. Temporal gaps in the marker data caused by occlusions were then manually filled within the Vicon Nexus software to reconstruct the full-body motion model.

The RGB/RGBD cameras operated at a frame rate of 30fps, while the Vicon motion capture system recorded marker trajectories at 100Hz. Temporal synchronisation between the camera and the Vicon system was achieved by aligning their temporal references using a trigger-based event (a clap action), followed by timestamp alignment and interpolation. {Further details of the synchronisation procedure are provided in Appendix~B.}

%\mmn{I think this synchronisation part is not within the direct scope of the paper and so should go into an Appendix.} \jj{has removed to appendix now}

\subsubsection{Spatial alignment of skeleton data}

We calibrated the cameras' intrinsic parameters and the extrinsic parameters between each camera and the Vicon system, respectively. See Appendix C for details.

To enable a consistent comparison between camera-based pose estimation and motion capture data, joints defined in the Vicon marker-based template are mapped to COCO-style skeleton joints \cite{lin2014microsoft}, as  illustrated in Fig.~\ref{fig:template_RTMO} and summarised in Table~\ref{alignment}.
In particular, the left and right hip joint centers are not directly available in the mocap data. Following the approach described in \cite{bell1990comparison}, the hip joint locations are estimated using the posterior and anterior superior iliac spine markers ($PSIS_R$, $PSIS_L$, $ASIS_R$, $ASIS_L$) provided by the Vicon system. In Table~\ref{alignment}, $mid(x,y)$ denotes the midpoint between joints $x$ and $y$. This joint alignment procedure ensures anatomical consistency between the camera-based skeleton and the Vicon reference skeleton.

\begin{figure}[t]
\centering
\begin{minipage}[t]{0.42\columnwidth}
    \vspace{0pt}
    \centering
    \includegraphics[width=\linewidth]{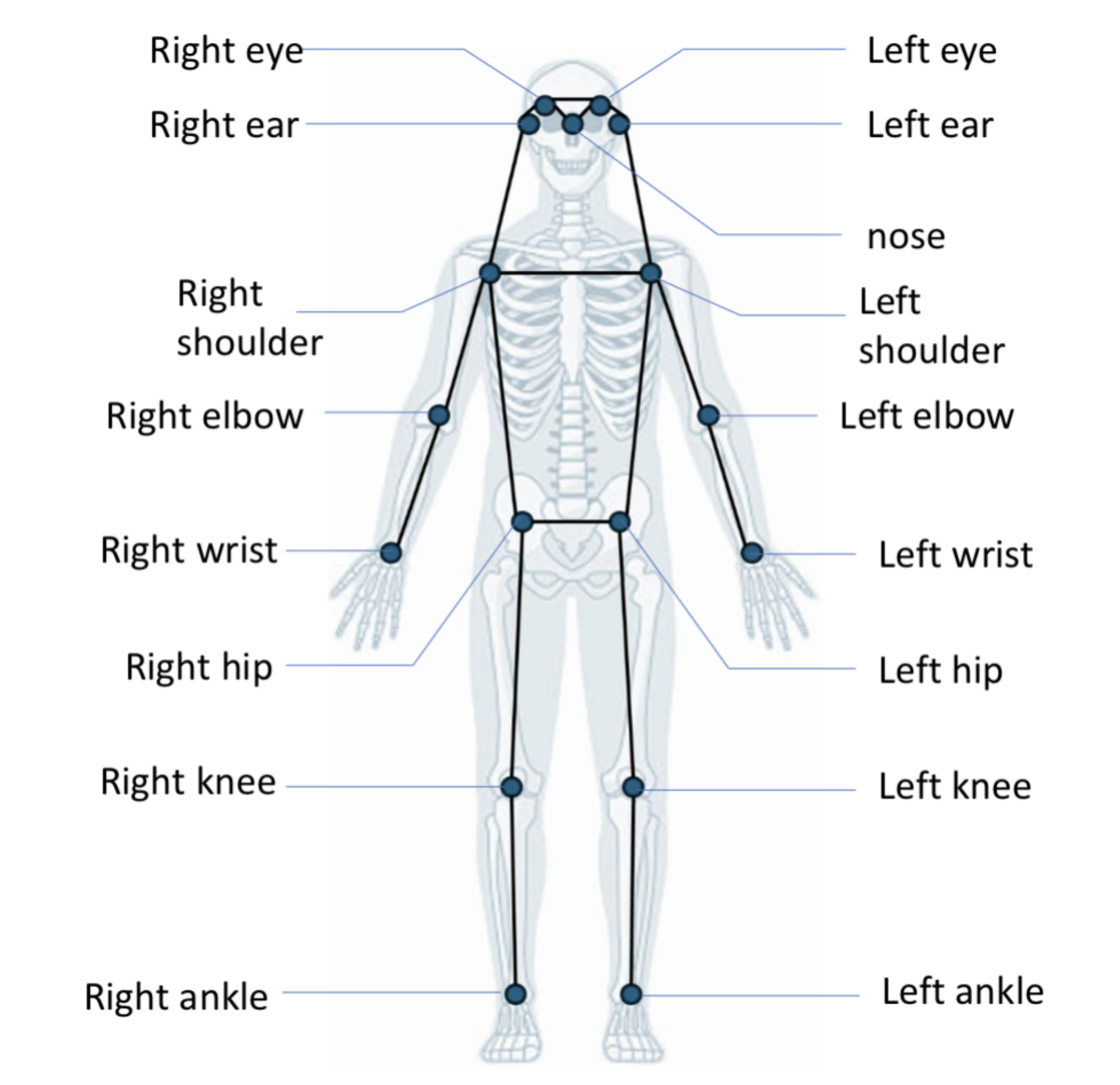}
    \captionof{figure}{COCO style 2D human pose representation \cite{lin2014microsoft}.}
    \label{fig:template_RTMO}
\end{minipage}
\hfill
\begin{minipage}[t]{0.50\columnwidth}
    \vspace{30pt}
    \centering
    \footnotesize
    \captionof{table}{Human skeleton joints alignment across templates.}
    \label{alignment}

    \resizebox{\linewidth}{!}{
    \begin{tabular}{c|c}
    \toprule
    COCO style Pose \cite{lin2014microsoft} & Pose template in Vicon system \\
    \hline
    Left shoulder & $SHO\_L$\\\hline
    Right shoulder & $SHO\_R$\\\hline
    Left elbow & $mid(ELB\_LAT\_L, ELB\_MID\_L)$\\\hline
    Right elbow & $mid(ELB\_LAT\_R, ELB\_MID\_R)$\\\hline
    Left wrist & $mid(WRI\_LAT\_L, WRI\_MID\_L)$\\\hline
    Right wrist & $mid(WRI\_LAT\_R, WRI\_MID\_R)$\\\hline
    Left hip & Bell et al. \cite{bell1990comparison} ($PSIS\_R, PSIS\_L, ASIS\_R, ASIS\_L$)\\\hline
    Right hip & Bell et al. \cite{bell1990comparison} ($PSIS\_R, PSIS\_L, ASIS\_R, ASIS\_L$)\\\hline
    Left knee & $mid(KNEE\_LAT\_L, KNEE\_MID\_L)$\\\hline
    Right knee & $mid(KNEE\_LAT\_R, KNEE\_MID\_R)$\\\hline
    Left ankle & $mid(ANK\_LAT\_L, ANK\_MID\_L)$\\\hline
    Right ankle & $mid(ANK\_LAT\_R, ANK\_MID\_R)$\\
    \bottomrule
    \end{tabular}
    }
\end{minipage}
\end{figure}

\begin{comment}
\begin{figure}[t]
    \centering
        \includegraphics[width=0.4\columnwidth]{images/coco.pdf}
    \caption{COCO style 2D human pose representation \cite{lin2014microsoft}.}
\label{fig:template_RTMO}
\end{figure}
\end{comment}
\begin{comment}
\begin{table}[htbp]
\centering
\footnotesize
\caption{Human skeleton joints alignment across templates.}
\resizebox{0.6\columnwidth}{!}{
\begin{tabular}{c|c}
\toprule
COCO style Pose \cite{lin2014microsoft} & Pose template in Vicon system \\ \hline

Left shoulder & $SHO\_L$\\\hline

Right shoulder & $SHO\_R$\\\hline

Left elbow & $mid(ELB\_LAT\_L, ELB\_MID\_L)$\\\hline

Right elbow & $mid(ELB\_LAT\_R, ELB\_MID\_R)$\\\hline

Left wrist & $mid(WRI\_LAT\_L, WRI\_MID\_L)$\\\hline

Right wrist & $mid(WRI\_LAT\_R, WRI\_MID\_R)$\\\hline

Left hip & Bell et al. \cite{bell1990comparison} ($PSIS\_R, PSIS\_L, ASIS\_R, ASIS\_L$)\\\hline

Right hip & Bell et al. \cite{bell1990comparison} ($PSIS\_R, PSIS\_L, ASIS\_R, ASIS\_L$)\\\hline

Left knee & $mid(KNEE\_LAT\_L, KNEE\_MID\_L)$\\\hline

Right knee & $mid(KNEE\_LAT\_R, KNEE\_MID\_R)$\\\hline

Left ankle & $mid(ANK\_LAT\_L, ANK\_MID\_L)$\\\hline

Right ankle & $mid(ANK\_LAT\_R, ANK\_MID\_R)$\\\bottomrule
\end{tabular}
}
\label{alignment}
\end{table}
\end{comment}

\subsubsection{Human pose evaluation}
\label{Sec:human_pose_ev}
%{For the camera-based data, 2D human joint positions are first extracted using the RTMO pose estimation method \cite{lu2024rtmo}. For RGBD cameras, the 2D keypoints are further combined with the corresponding depth measurements to recover 3D joint positions. To improve robustness against depth noise, the depth value for each joint is computed as the median within a 3×3 neighbourhood around the corresponding 2D keypoint. In contrast, RGB cameras provide only 2D pose estimates due to the lack of depth sensing.}

For the camera-based data, 2D human joint positions are extracted independently from the same recordings using recent pose estimation methods. {We consider several candidate estimators, including MediaPipe~\cite{lugaresi2019mediapipe} and AlphaPose~\cite{fang2022alphapose}, and selected RTMO~\cite{lu2024rtmo} and YOLO26~\cite{sapkota2025yolo26}, which gave the strongest overall performance among the real-time methods in preliminary testing.} These two estimators are state-of-the-art, single-stage, real-time, and optimised for efficient inference on edge hardware (e.g., NVIDIA Jetson platforms), making them well suited for continuous in-home healthcare monitoring. {{Note that they differ in their underlying design:} RTMO is a bottom-up one-stage method, whereas YOLO26 couples person detection and keypoint regression in a single end-to-end network.}

% , so agreement between them indicates that the camera-level differences we report are not specific to a single architecture.

% While several other methods, such as MediaPipe~\cite{lugaresi2019mediapipe} and AlphaPose~\cite{fang2022alphapose}, were also considered, RTMO and YOLO26 were selected as they provided the most} accurate and computationally efficient pose estimation in our preliminary evaluations.
%For the camera-based data, 2D human joint positions are extracted using the RTMO pose estimation method \cite{lu2024rtmo}. \mm{While several other methods, such as Alphapose, Mediapipe etc. (CITE) were carefully considered,}RTMO was selected because \mmn{in our preparatory investigations it provided the most} accurate and computationally efficient pose estimation and ran efficiently on Jetson \mmn{again, not sure mentioning Jetsons is necessary unless you have good reason.}
% unnecessary: , making it suitable for evaluating camera-based human pose sensing without introducing significant computational overhead.} 

{The same depth-lifting procedure described below was applied identically to the 2D keypoints produced by both estimators.} For RGBD cameras, the estimated 2D keypoints are first transformed from the image coordinate system to the camera coordinate system using the intrinsic calibration parameters. Depth information is then incorporated to recover the corresponding 3D joint positions. To improve robustness against depth noise and local outliers, the depth value associated with each joint is computed as the median depth within a $3 \times 3$ pixel neighborhood centered at the corresponding 2D keypoint location. {In contrast, for RGB cameras, only 2D joint positions are estimated, as no depth information is available to recover 3D joint coordinates.} 
%\mmn{How do we estimate 3D joints for the RGB camera?}

Reference 3D joint data are collected using the motion capture system and treated as the ground-truth measurements. The 3D joint positions are transformed into the camera coordinate system using the extrinsic parameters.

Following existing studies in human pose estimation, multiple quantitative metrics are adopted to evaluate both 2D and 3D pose estimation performance. For 2D pose evaluation, OKS-based mean average precision (mAP) and percentage of correct keypoints (PCK) are used \cite{andriluka20142d,lin2014microsoft}.
MAP measures the overlap between predicted and ground-truth keypoints, normalized by the person's scale, and is calculated across multiple OKS thresholds (from 0.5 to 0.95, at steps of 0.05) to reflect overall localization precision. PCK measures the fraction of predicted joints that fall within a specific error margin. We use PCK@0.2, which represents the percentage of predicted keypoints that fall within a matching threshold of $0.2 \times L_{norm}$. $L_{norm}$ is defined as the distance between the left shoulder and right hip, ensuring the metric remains invariant to the subject's scale within the image.
For 3D pose evaluation, 
mean per-joint position error (MPJPE) provides a direct measure of error by calculating the average Euclidean distance for all joints between the predicted and ground-truth human pose, and 
3D PCK reports the fraction of joints where the error is below a fixed threshold (e.g., $150mm$), offering a measure of the model's robustness against significant depth or localization outliers \cite{ionescu2013human3,mehta2017monocular}. All metrics are computed separately for each pose estimator}.

\section{Results}
We provide comprehensive evaluation results of the proposed validation {protocols} across both sensor-level and application-level perspectives. Specifically, we first analyse the metrological performance of the evaluated cameras, including depth measurement bias and precision, temperature and stability during continuous operation, and field-of-view characteristics. These measurements quantify the fundamental sensing properties of the devices under different environmental conditions. We then evaluate the performance on downstream pose estimation tasks using both 2D and 3D evaluation metrics.
For 2D pose estimation, mAP and PCK are reported, while for 3D pose estimation we report MPJPE and PCK.

\subsection{Metrological Performance Evaluation}
\label{Sec:metrological}
\begin{figure*}[t]
    \centering
    \includegraphics[width=0.32\columnwidth]{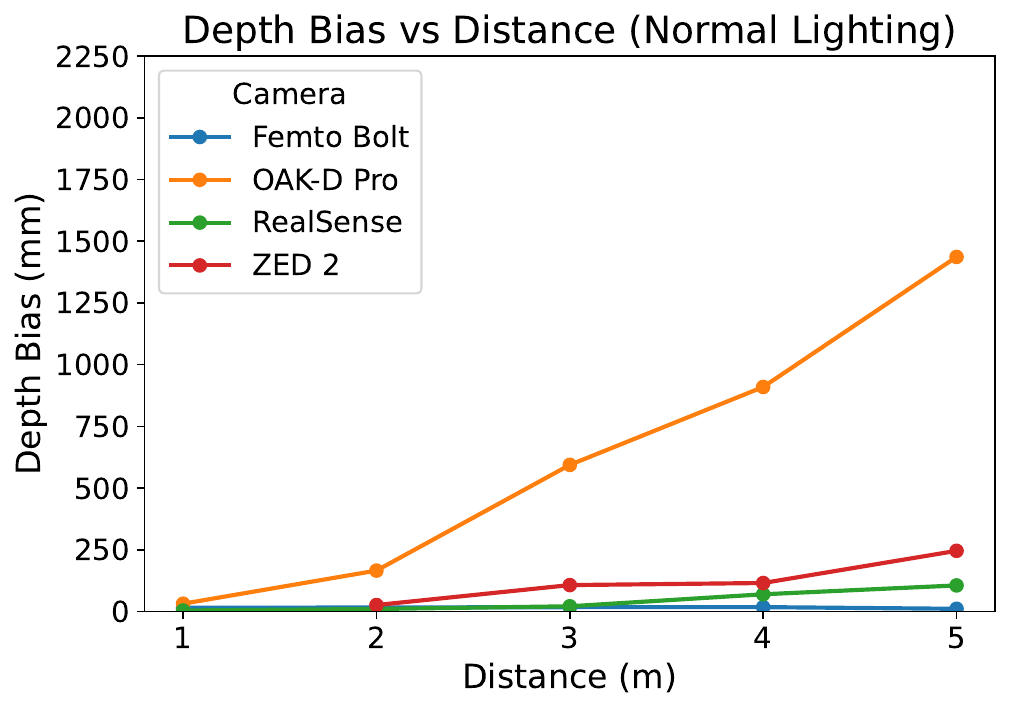}
    \includegraphics[width=0.32\columnwidth]{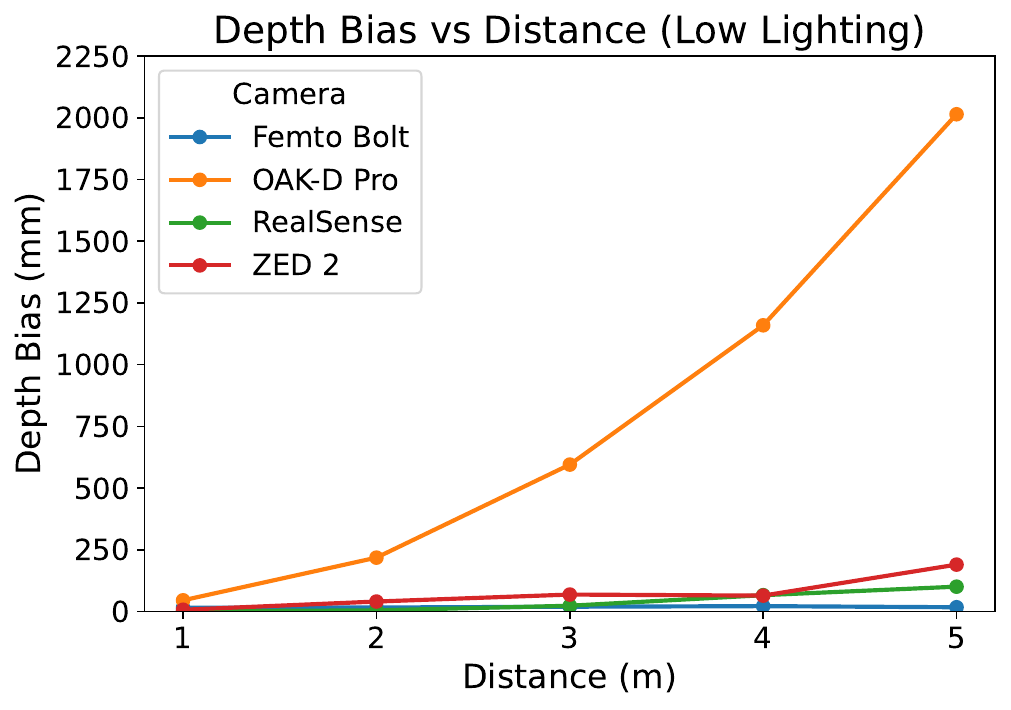}
    \includegraphics[width=0.32\columnwidth]{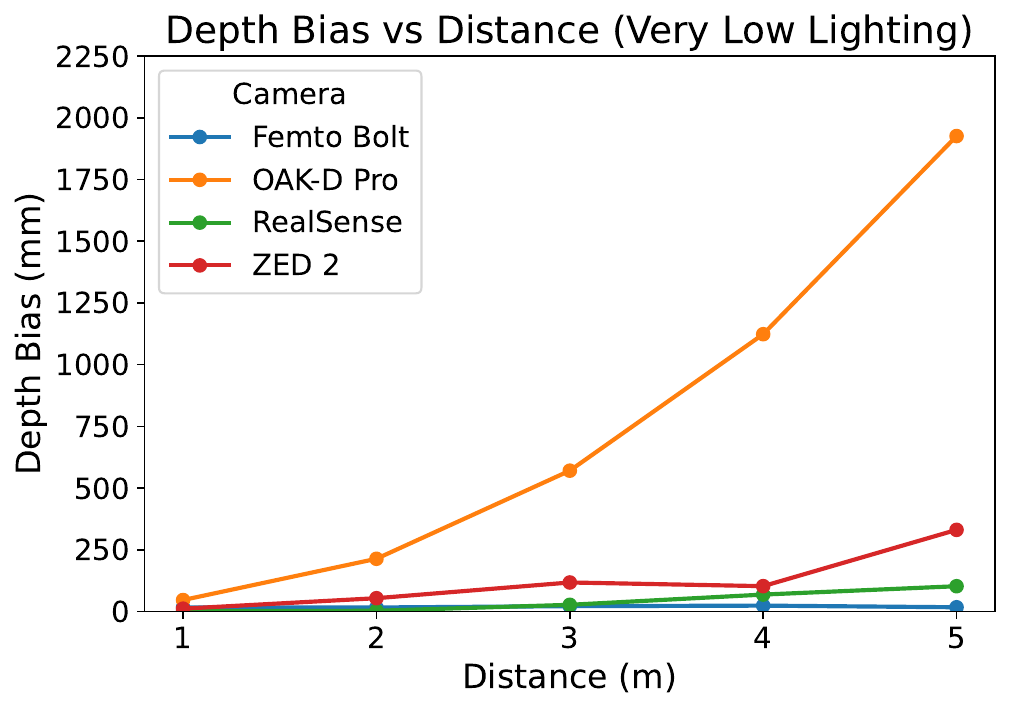}
    \caption{Bias of the tested cameras at different camera-to-target distances under different lighting condition.}
    \label{fig:depth_bias}
\end{figure*}

\noindent{\bf Depth measurement performance}
Fig.~\ref{fig:depth_bias} illustrates the depth bias 
%\amn{Fig 8, 9 talk about bias and percision etc but they are not defined in the paper how did we calculate them. }{The evaluation metrics should be stated in the method section as this is part of method to say what we want to measure. I also put the evaluation metrics of exercise 2 in the method part.} \amn{maybe refer them to the method sec e.g. (eq. ?). } 
as a function of camera-to-target distance under normal-light, low-light, and very low-light lighting conditions for four RGB-D cameras. Across all lighting conditions, {while a distance-dependent increase in depth bias is expected, all devices exhibit this trend, except Femto Bolt which exhibits a markedly different behaviour, with its depth bias remaining nearly constant across the tested distance range, varying only within a narrow range (e.g., within $11.17-17.98 mm$ under normal lighting, $15.85-21.99mm$ under low lighting and $16.52-24.06mm$ under very low lighting).}
%\mmn{this is I guess quite obvious, so rather than saying that, can we just say "while this is expected, the cameras behave differently as outlined next".}.

Among the evaluated cameras, OAK-D Pro consistently exhibits the largest bias, with a pronounced monotonic increase as distance grows. This effect is particularly evident beyond $3m$, where the bias rises sharply under all lighting conditions and becomes more severe in low and very low lighting environments. In contrast, Femto Bolt shows the smallest bias across the entire distance range, maintaining relatively stable and low error levels even at longer distances. When averaged over all distances, the bias of Femto Bolt is $15.77mm$ under normal-light conditions, $18.27mm$ under low-light conditions, and $19.29mm$ under very low-light conditions. For RealSense, the depth bias remains very small and comparable to that of Femto Bolt at distances up to $3m$ across all lighting conditions. A noticeable increase in bias is only observed beyond $3m$. However, the magnitude of this increase remains relatively limited compared to OAK-D Pro. {ZED 2 shows a moderate increase in bias with distance that lies between that of the OAK-D Pro and the Intel RealSense, along with a noticeable sensitivity to reduced illumination.}

\begin{figure*}[t]
    \centering
    \includegraphics[width=0.32\columnwidth]{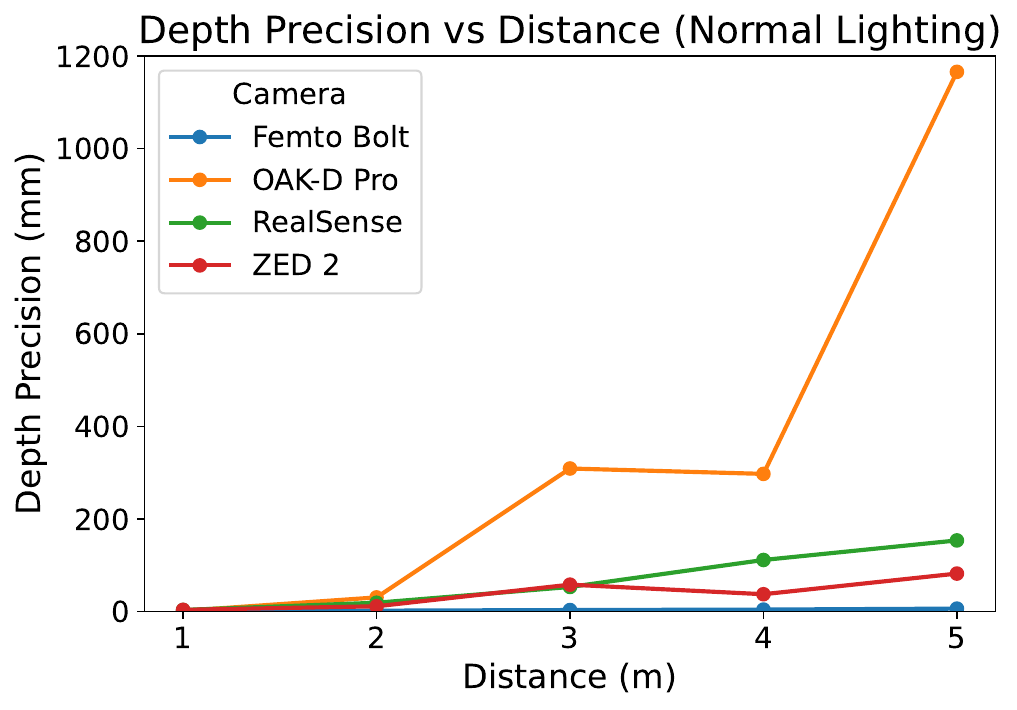}
    \includegraphics[width=0.32\columnwidth]{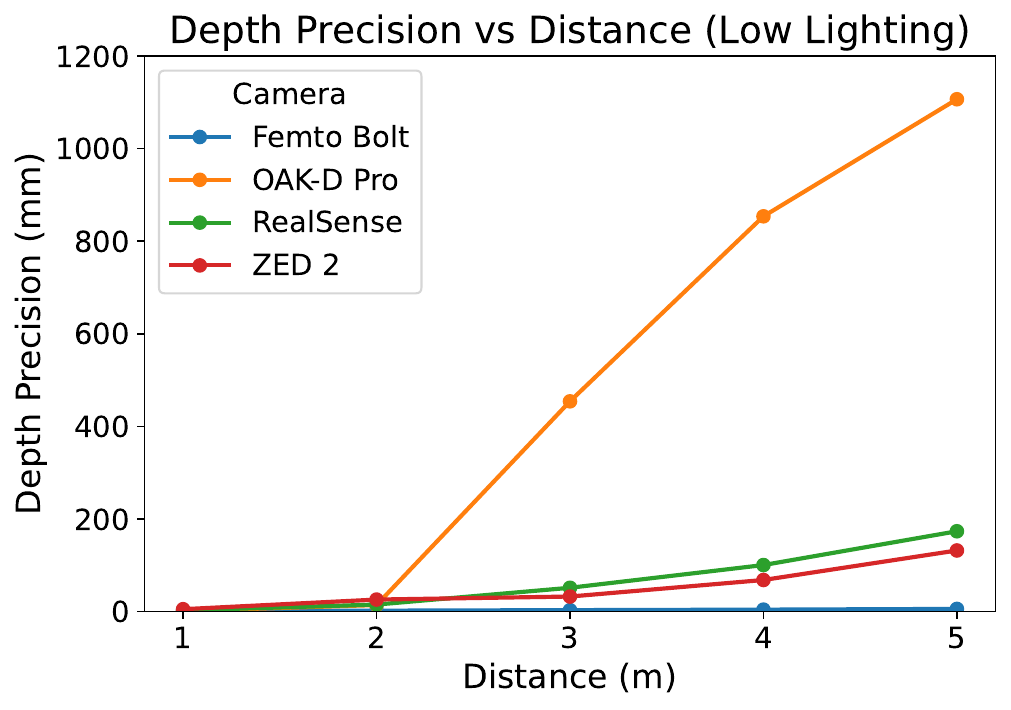}
    \includegraphics[width=0.32\columnwidth]{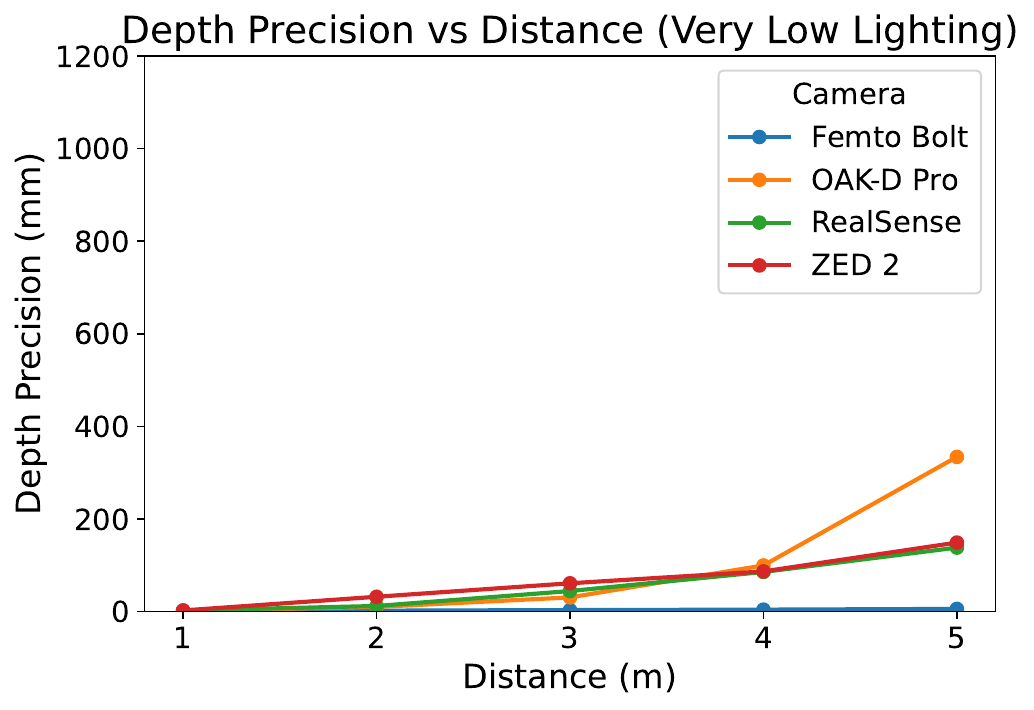}
    \caption{Precision of the tested cameras at different camera-to-target distances under different lighting condition. }
    \label{fig:depth_precision}
\end{figure*}

Fig.~\ref{fig:depth_precision} presents the depth precision as a function of camera-to-target distance under normal, low, and very low lighting conditions for the four evaluated RGB-D cameras. Overall, {as expected,} depth precision exhibits a clear dependence on measurement distance, with variability increasing as the distance grows, particularly under reduced illumination. {However, some interesting observations can be made.}
%This trend is broadly consistent with the behaviour observed for depth bias.  TOO VAGUE
Across all lighting conditions, Femto Bolt demonstrates the most stable precision, maintaining consistently low variability over the entire distance range, even at longer distances. In contrast, OAK-D Pro shows a pronounced degradation in precision as distance increases, with a sharp rise beyond $3m$. At a distance of $5m$, the precision of OAK-D Pro is $1165.7mm$ under normal-light conditions, $1106.6mm$ under low-light conditions, and $334.1mm$ under very low-light conditions. The improved precision under very low-light conditions may be attributed to reduced interference from ambient illumination, leading to more stable active depth sensing. RealSense and ZED 2 generally exhibit precision values that lie between those of the Femto Bolt and OAK-D Pro. {Under very low lighting conditions or at short distances, however, the precision of OAK-D Pro becomes comparable to that of RealSense and ZED 2.}

\begin{figure*}[t]
    \centering
    \includegraphics[width=0.32\columnwidth]{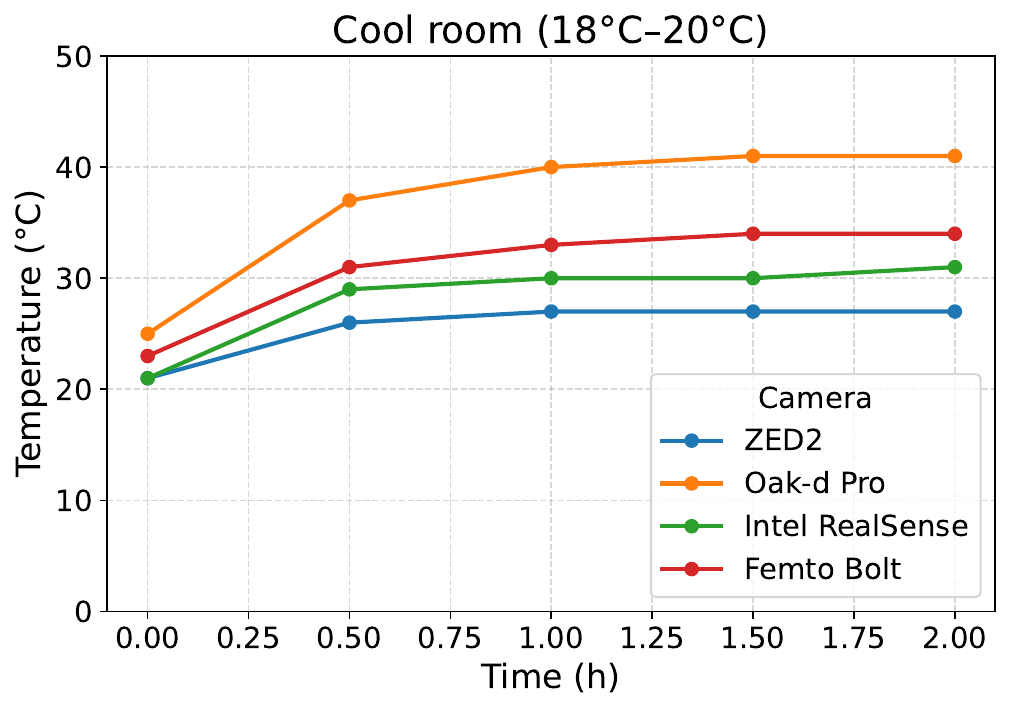}
    \includegraphics[width=0.32\columnwidth]{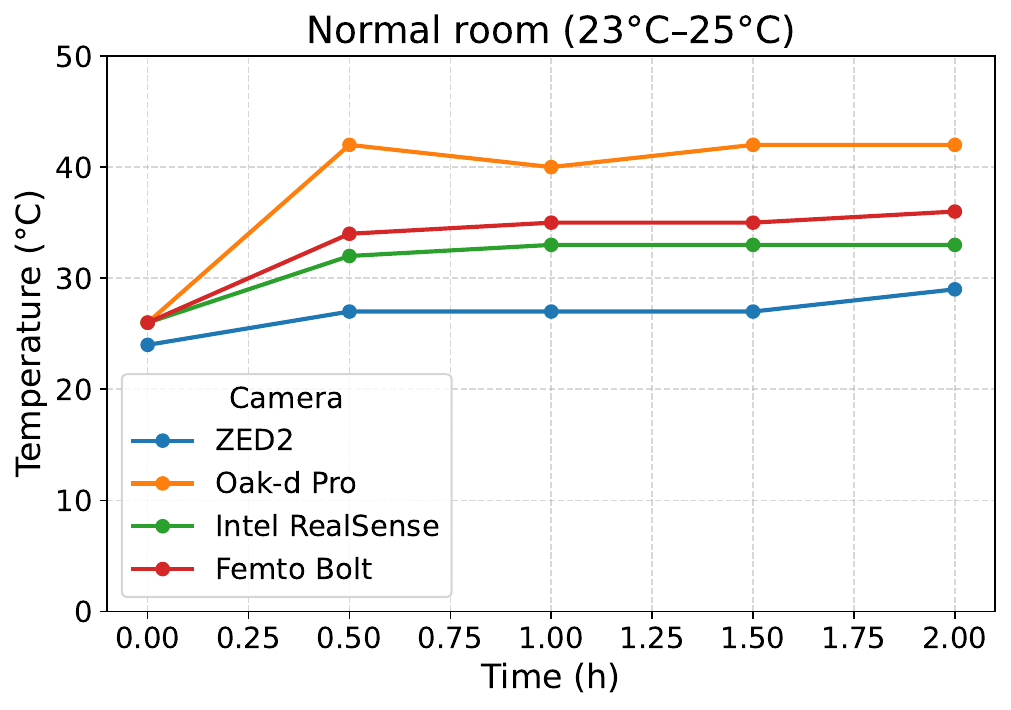}
    \includegraphics[width=0.32\columnwidth]{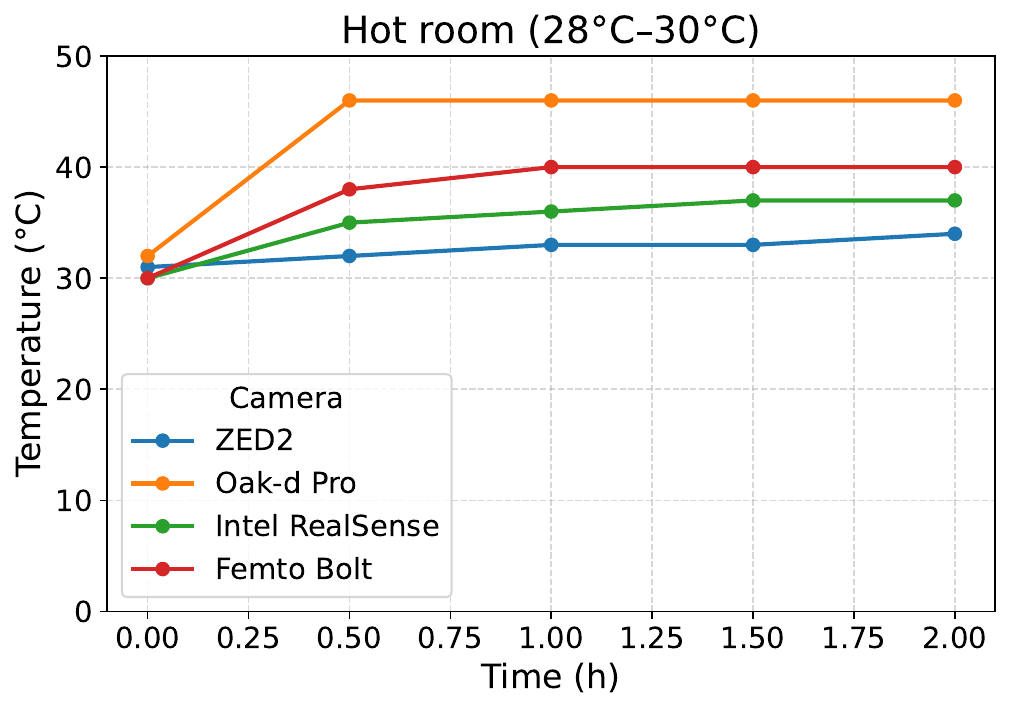}
    \caption{Temperatures of tested cameras after running for 2h in different room temperatures.}
    \label{fig:temperature}
\end{figure*}

\vspace*{3mm}

\noindent {\bf Temperature and stability performance --}
Temperature is evaluated for the four RGB-D cameras under two influencing factors: different operating modes and ambient room temperatures. To examine the impact of ambient temperature, the cameras {are operated for two hours under three room temperature conditions}: cool (18--20$^\circ$C), normal (21--24$^\circ$C), and warm (25--28$^\circ$C). During this period, the device temperature variations and depth measurement drift are recorded to assess system stability.
As shown in Fig.~\ref{fig:temperature}, across all ambient temperatures, a rapid temperature increase is observed within the first half-hour, followed by a gradual stabilisation phase. OAK-D Pro consistently reaches the highest operating temperature, exhibiting the largest rise across all room conditions. After 2 hours of operation, the temperature of the OAK-D Pro reaches 41$^\circ$C in a cool room, 42$^\circ$C in a normal room, and 46$^\circ$C in a hot room.
Femto Bolt and RealSense show moderate temperature increases, stabilising at very low absolute temperatures, while ZED2 maintains the lowest and most stable operating temperature. Higher ambient room temperatures lead to elevated steady-state device temperatures for all cameras, although the relative ordering between devices remains consistent across conditions. 

\begin{figure}[t]
    \centering
    \includegraphics[width=0.9\columnwidth]{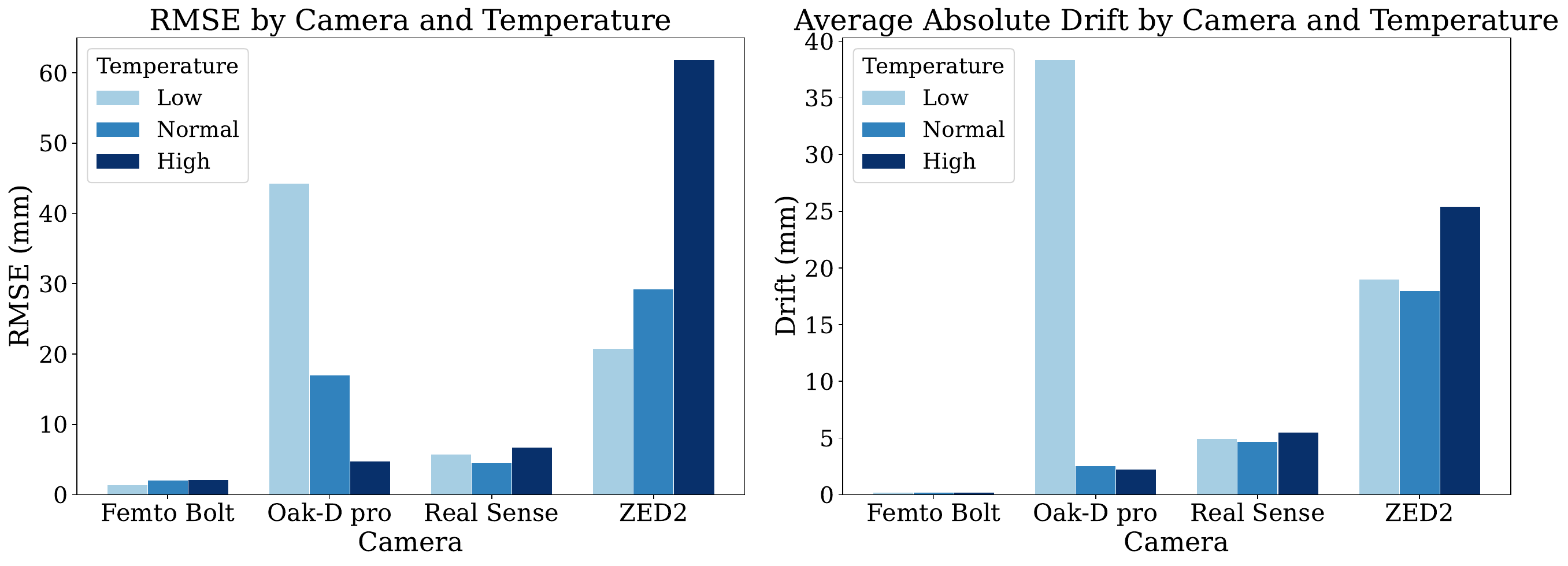}
    \caption{The stability of tested cameras after running for 2h in different room temperatures. }
    \label{fig:stability}
\end{figure}

Fig.~\ref{fig:stability} presents the depth measurement stability of the cameras evaluated at low, normal, and high room temperatures by RMSE and AAD. Femto Bolt consistently exhibits the lowest RMSE and drift across all temperature conditions, indicating high stability with minimal temperature-induced variation. In contrast, OAK-D Pro and ZED 2 exhibit substantial increases in both RMSE and drift as temperature rises, suggesting strong sensitivity to temperature variation. RealSense generally shows moderate changes, with RMSE and drift values relatively higher than those of the Femto Bolt. %, but not reaching the highest values observed among the evaluated cameras.}

\vspace*{3mm}

\noindent {\bf Field of view performance --}
%\amn{ the Sony See3CAM and Logitech BRIO (if it is the logitec please remove brio) appear in FOV \amn{acronym needs to be define first...} but not elsewhere - this needs to be addressed in whole of the paper.} \jj{Cameras need to be specified in the method section.}
For FOV assessment, we tested multiple resolution settings to analyze how they affect the field of view for each camera. The measured FOV values were compared against the manufacturer's reported specifications to calculate residuals and verify accuracy, with comparison results given in Table~\ref{tab:fov}.

\begin{figure}[H]
    \centering
        \includegraphics[width=1.0\columnwidth]{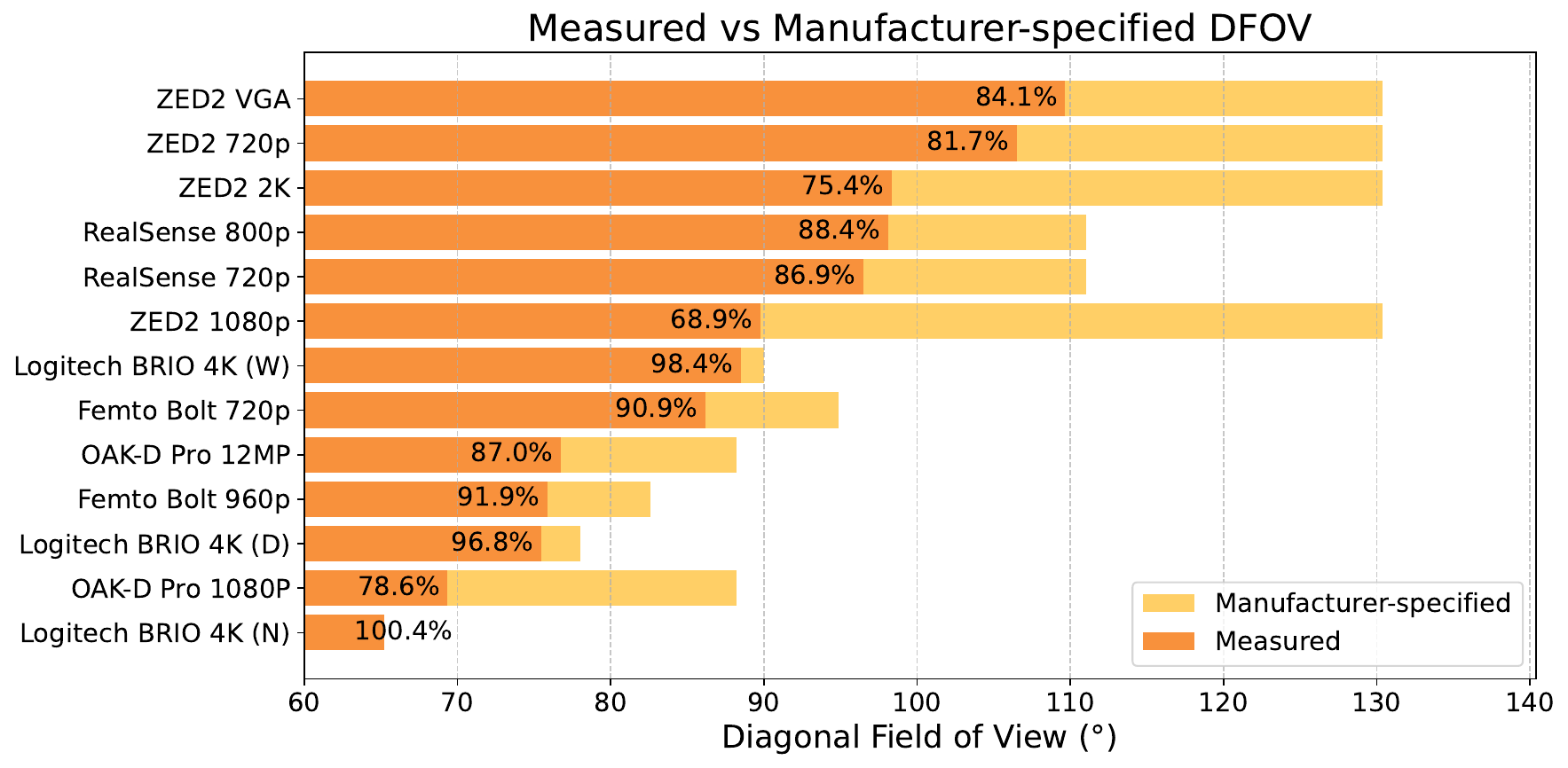}
    \caption{Comparison of measured and manufacturer-specified DFOV across multiple camera models and resolutions. Percentage annotations show how each measurement compares to the corresponding manufacturer reported. D: Diagonal, W: Widest, N: Narrowest.}
\label{fig:DFOV}
\end{figure}

\begin{table}[t]
\footnotesize
\centering
\caption{Comparison between measured and manufacturer-specified FOV across cameras and resolution settings.}\label{tab:fov}
\begin{tabular}{llccccc}
\hline
\textbf{Camera} & \textbf{Setting} & \multicolumn{3}{c}{\textbf{Measured FOV (°)}} & \multicolumn{2}{c}{\textbf{Specified FOV (°)}} \\
 & & \textbf{HFOV} & \textbf{VFOV} & \textbf{DFOV} & \textbf{HFOV} & \textbf{VFOV} \\ \hline
ZED2       & 2K: 2208$\times$1242   & 90.55 & 59.11 & 98.37  & 110 & 70 \\
ZED2       & 1920$\times$1080       & 81.88 & 52.25 & 89.80  & 110 & 70 \\
ZED2       & 1280$\times$720        & 99.39 & 64.89 & 106.51 & 110 & 70 \\
ZED2       & VGA 672$\times$376     & 102.00 & 69.96 & 109.67 & 110 & 70 \\ \hline
OAK-D Pro  & 1080P                  & 62.21 & 37.45 & 69.37  & 69  & 55 \\
OAK-D Pro  & 12MP: 4056$\times$3040 & 64.88 & 50.62 & 76.77  & 69  & 55 \\ \hline
Femto Bolt & 1280$\times$720        & 78.10 & 49.98 & 86.19  & 80  & 51 \\
Femto Bolt & 1280$\times$960        & 63.94 & 50.13 & 75.90  & 65  & 51 \\ \hline
RealSense  & 1280$\times$720        & 88.65 & 57.59 & 96.52  & 90  & 65 \\
RealSense  & 1280$\times$800        & 88.65 & 62.92 & 98.11  & 90  & 65 \\ \hline
Logitech BRIO 4K & 1280$\times$720 (D) & -- & -- & 75.49 & \multicolumn{2}{c}{78\textsuperscript{\scriptsize D}} \\
Logitech BRIO 4K & 1280$\times$720 (N) & -- & -- & 65.23 & \multicolumn{2}{c}{65\textsuperscript{\scriptsize D}} \\
Logitech BRIO 4K & 1280$\times$720 (W) & -- & -- & 88.52 & \multicolumn{2}{c}{90\textsuperscript{\scriptsize D}} \\ \hline
\multicolumn{7}{l}{\scriptsize\textsuperscript{D} Manufacturer reports diagonal FOV only.} \\
\end{tabular}
\end{table}
From the FOV evaluations, we observe that most cameras showed slight variations from manufacturer specifications, with the ZED2 showing the largest deviation at some resolutions 
%\mmn{you should reproduce the manuf. numbers otherwise this comparison is not so meaningful.}.  
The Femto Bolt and RealSense cameras demonstrated the most consistent and accurate FOV measurements relative to their specifications across different resolutions.  
Resolution settings significantly affect FOV in most cameras, with some showing considerable variations between different resolution settings. To ensure a fair and aspect-ratio-independent comparison across cameras, we further assess the diagonal field of view (DFOV) for each device. The quantitative results are provided in the 3\textsuperscript{rd} column of Table~\ref{tab:fov}, and Fig.~\ref{fig:DFOV} presents a visual comparison between the measured DFOV and manufacturer specifications. More detailed comparisons of the cameras’ Horizontal Field of View (HFOV) and Vertical Field of View (VFOV) with their corresponding manufacturer specifications are presented in  Appendix~\ref{appendix_fov}.

{When comparing all cameras at the standard resolution of 1280$\times$720, the field of view ranking from widest to narrowest is given in Table~\ref{tab:fov_res}. ZED2 exhibits the largest measured DFOV, followed by the RealSense and Logitech BRIO 4K, while the Femto Bolt and OAK-D Pro provide comparatively narrower fields of view.}

\begin{table}[H]
\footnotesize
\centering
\caption{Camera field-of-view comparison at $1280\times720$ resolution. Cameras are ranked by diagonal field of view (DFOV) from widest to narrowest.}
\begin{tabular}{ccccc}
\hline
\textbf{Rank} & \textbf{Camera} & \textbf{HFOV (°)} & \textbf{VFOV (°)} & \textbf{DFOV (°)} \\ \hline
1 & ZED2             & 99.39 & 64.89 & 106.52 \\
2 & RealSense        & 88.65 & 57.59 & 96.62  \\
3 & Logitech BRIO 4K\footnotemark & --    & --    & 88.52  \\
4 & Femto Bolt       & 78.10 & 49.98 & 86.19  \\
5 & OAK-D Pro        & 62.21 & 37.45 & 69.37  \\ \hline
\end{tabular}
\label{tab:fov_res}
\end{table}

\footnotetext{Entry shows ``-- × --, 88.52 (D)", where dashes denote unavailable HFOV and VFOV values. The manufacturer of the Logitech BRIO 4K reports only the diagonal field of view. Therefore, only DFOV is measured in this work.}

\vspace*{3mm}

\begin{figure}[t]
    \centering
    \includegraphics[width=0.7\columnwidth]{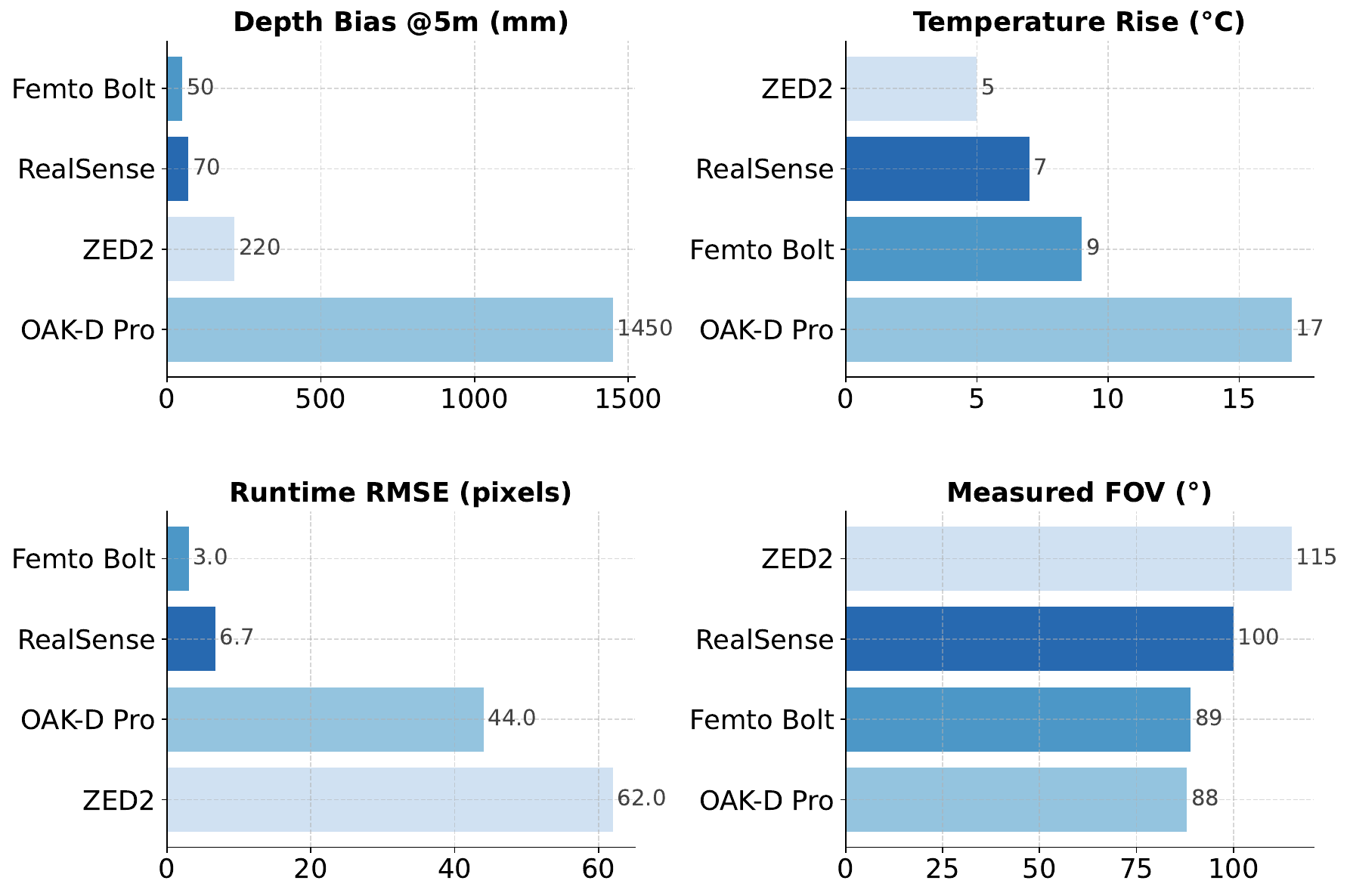}
    \caption{Comparative evaluation of the four depth cameras across key performance metrics. For depth bias, temperature rise, and runtime RMSE, very low values indicate better performance. In contrast, higher values are preferable for measured FOV. }
    \label{fig:four_measure_overview}
\end{figure}

\begin{figure}[t]
    \centering
    \includegraphics[width=0.5\columnwidth]{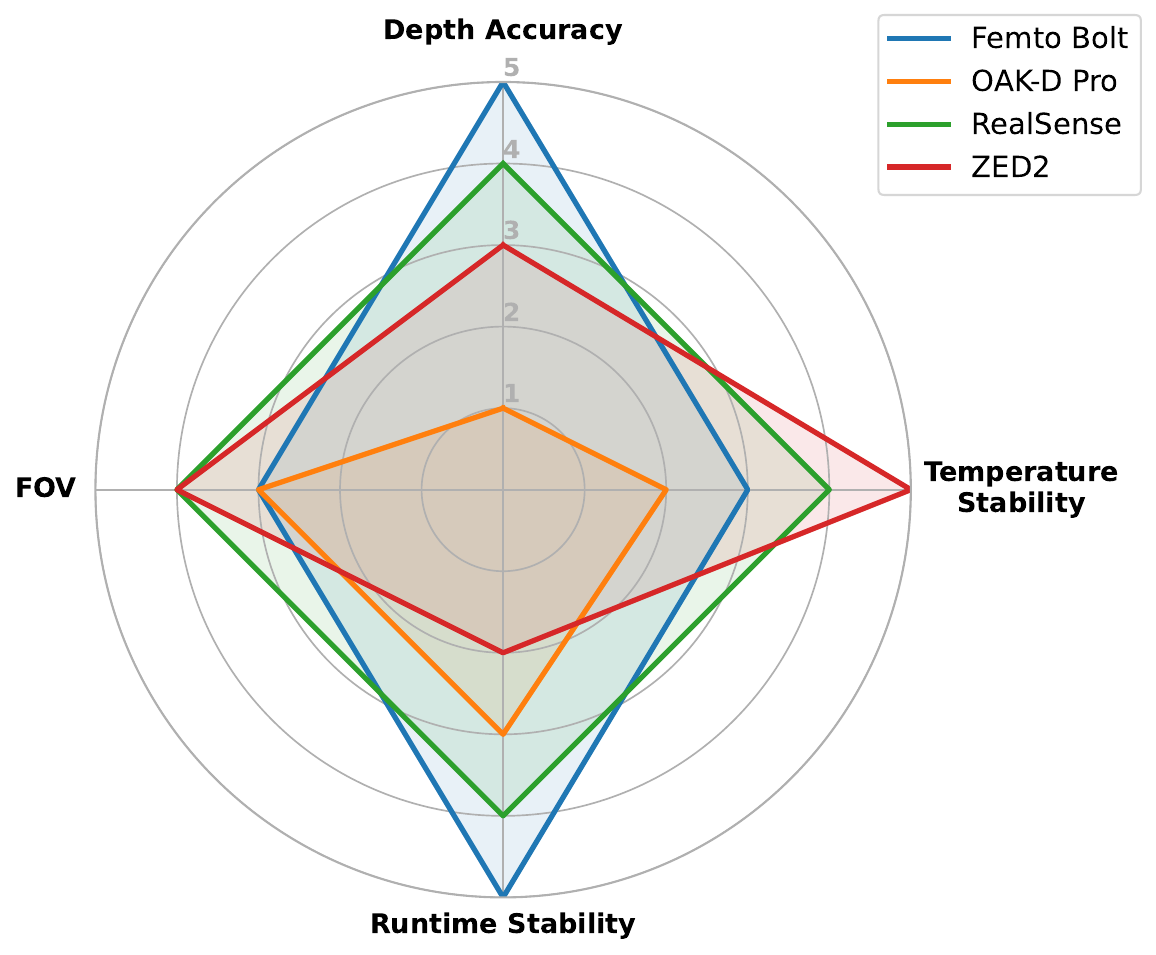}
    \caption{Overview of a qualitative score comparison for the four depth cameras. }
    \label{fig:depth_overview}
\end{figure}

\noindent {\bf Summary --} To facilitate a clear comparison, we combine the four key metrological evaluations, i.e., depth accuracy, temperature and stability, runtime, and FOV across all the four depth cameras. 
%A quantitative and qualitative summary is shown in Fig.~\ref{fig:four_measure_overview} and Fig.~\ref{fig:depth_overview}, respectively.
{Fig.~\ref{fig:four_measure_overview} presents a quantitative comparison of the cameras across the four metrics. For depth bias, temperature rise, and runtime RMSE, lower values indicate better performance, whereas a larger field of view is preferred. As shown in the figure, Femto Bolt consistently achieves the lowest depth bias and runtime RMSE, along with competitive thermal stability, while RealSense also demonstrates relatively low error and stable performance. In contrast, OAK-D Pro shows significantly higher depth bias and runtime RMSE, and ZED2 exhibits higher runtime error despite its advantage in FOV.}
Fig.~\ref{fig:depth_overview} further provides  a qualitative comparison, where the plotted values represent relative performance ranks rather than absolute measurement values. The overall evaluations suggest that Femto Bolt and RealSense achieve the most balanced performance across all metrics, whereas OAK-D Pro and ZED2 show trade-offs between accuracy, stability, and field of view.

\subsection{Accuracy in pose estimation}
Following established evaluation practice in human pose estimation \cite{lu2024rtmo}, OKS-based mean average precision (mAP) and percentage of correct keypoints (PCK) are used to assess 2D pose estimation performance. For 3D pose evaluation, mean per-joint position error (MPJPE) and PCK are employed. {All metrics are computed independently for both estimators RTMO and YOLO26 (see Section \ref{Sec:human_pose_ev})}. {As the protocol targets indoor healthcare monitoring where everyday objects may partially obstruct the camera's view \cite{mainsphere}}, performance metrics are computed for each combination of camera, lighting, and {occlusion} conditions, and then averaged using the arithmetic mean across three camera heights ($1.8m$, $2.0m$, and $2.2m$), selected to represent practical wall-mounted installation positions in typical residential environments where standard ceiling heights are approximately $2.4m$. As demonstrated by the field-of-view analysis in Section~\ref{Sec:Exp_data} and Table~\ref{tab:tilt_angle}, these heights ensure that a feasible tilt angle exists for most cameras to capture the full body of a standing subject. This aggregation strategy isolates the effects of lighting and occlusion from camera placement, which is examined separately in the height-specific analysis.

\begin{figure*}[t]
    \centering
    \includegraphics[width=0.99\linewidth]{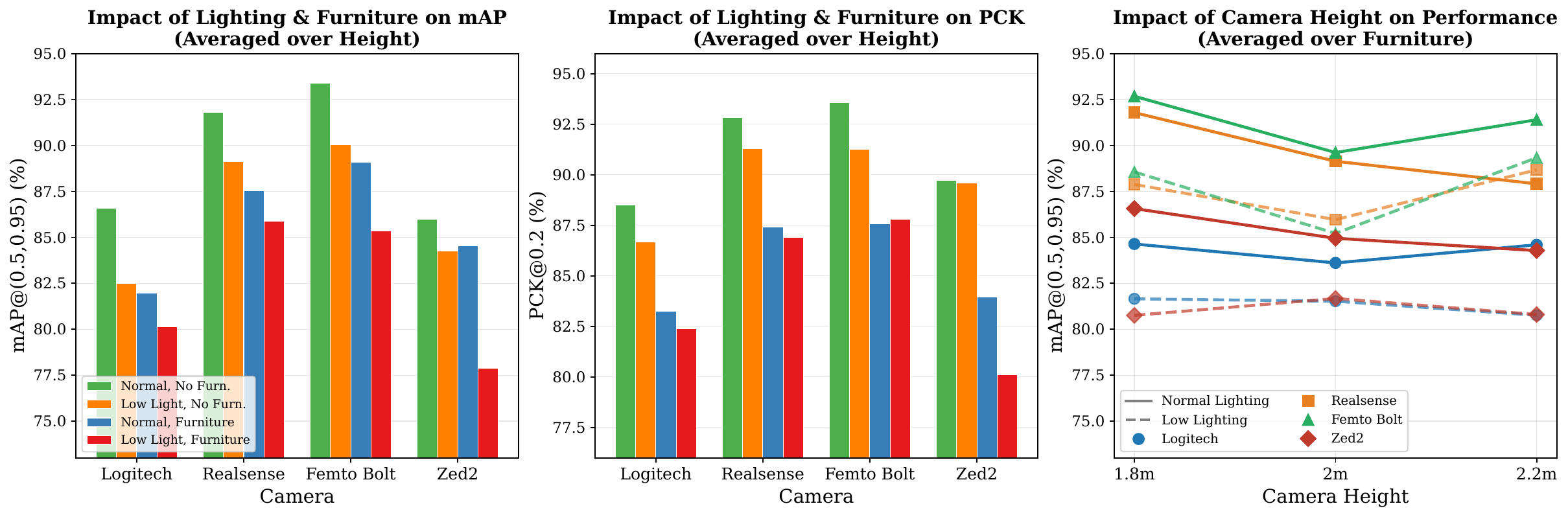}\\[2mm]
    {\small(\textbf{a}) RTMO}\\[3mm]
    \includegraphics[width=0.99\linewidth]{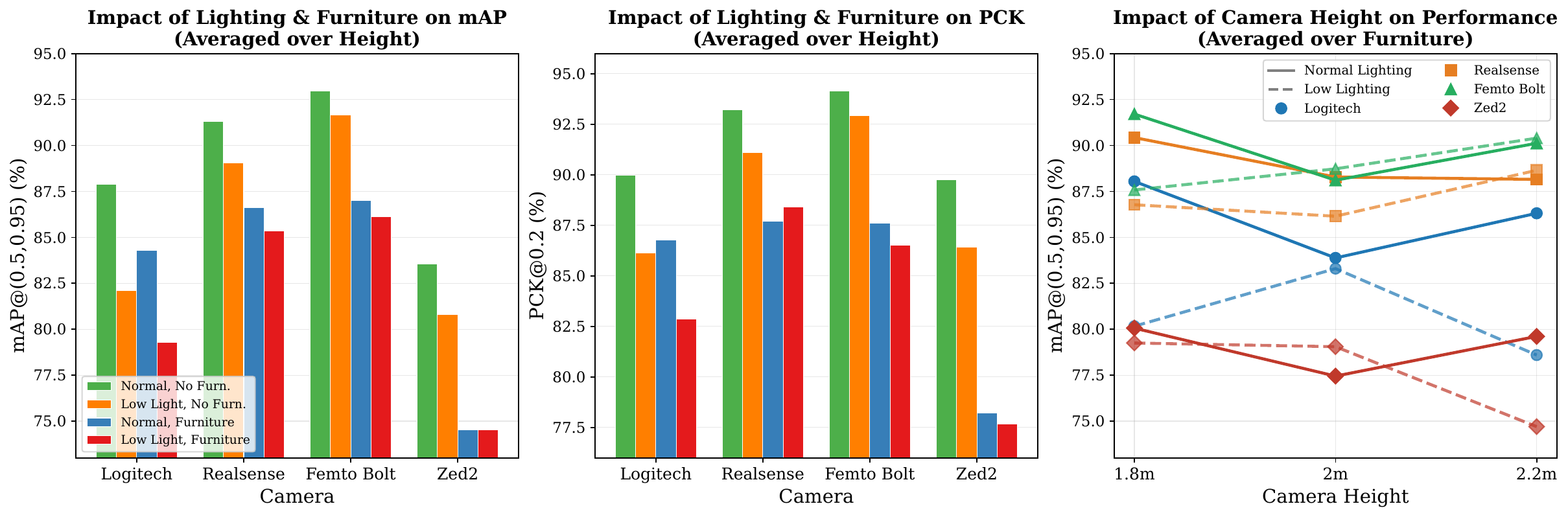}\\[2mm]
    {\small(\textbf{b}) YOLO26}
    \caption{{2D pose estimation performance across the four cameras for two pose
    estimators: (\textbf{a}) RTMO and (\textbf{b}) YOLO26. In each row, the left and
    centre panels show the impact of lighting and furniture occlusion on
    mAP@(0.5,0.95) and PCK@0.2 respectively (averaged over mounting height), and the
    right panel shows the impact of camera mounting height on mAP (averaged over
    furniture).}}
    \label{fig:2D_pose}
\end{figure*}

\vspace*{3mm}

\noindent {\bf 2D pose estimation performance --}
Fig.~\ref{fig:2D_pose} presents 2D pose estimation performance {obtained by applying RTMO and YOLO26 to the RGB streams captured by each camera,}
where across all four cameras, the highest mAP and PCK values are consistently observed under normal lighting, and without furniture, as expected. The two estimators produce a broadly consistent ranking across the camera inputs. With RTMO, the highest mean mAP is obtained on the Femto Bolt stream (89.5\%), followed closely by the RealSense stream (88.8\%). YOLO26 shows a similar trend, with mean mAP values of 89.4\% and 88.1\% for the Femto Bolt and RealSense streams, respectively. For both estimators, higher accuracy is generally achieved on the Femto Bolt and RealSense streams than on the Logitech and ZED2 streams across the evaluated conditions. Under the most challenging scenario (low lighting with furniture), both estimators show a clear degradation in performance across all camera inputs. With RTMO, mAP values decrease to \textasciitilde 85\%--86\% for the Femto Bolt and RealSense streams, and \textasciitilde 78\%--80\% for the Logitech and ZED2 streams. For YOLO26, the same pattern is observed, with mAP values of \textasciitilde 85\%--86\% for the Femto Bolt and RealSense streams and \textasciitilde 75\%--79\% for the Logitech and ZED2 streams. {For the Logitech stream, RTMO achieves performance comparable to the ZED2 stream, with mean mAP values of 82.8\% and 83.2\%, respectively.} {For YOLO26 however, the Logitech stream achieves a mean mAP of 83.4\% and clearly outperforms the ZED2 stream at a mean mAP of 78.4\%, indicating that YOLO26 is more sensitive than RTMO to the visual characteristics of the ZED2 RGB stream, potentially reflecting camera-specific properties such as output resolution, field of view, image sharpness, or noise under the evaluated indoor conditions.} 
%\mmn{If I was a referee, I would request to see all these numbers in a proper table - reading them off the bar charts in Fig 16 is rather difficult. The bar charts are nice and give a more broad overview...so perhaps add the table as well?}

% \mmn{this sounds misleading as if ZED is doing keypoint detection?} is somewhat more architecture-sensitive than on the other camera streams, while the RGB-only Logitech stream remains competitive in both cases.} \mmn{Can this last whole pgh be written such that the focus is on what the pose estimators achieve given the cameras and their specific hardware/resolution etc, rather than saying Femto Bolt achieves or ZED2 performs etc.}

{In summary, reduced illumination leads to a measurable decline in 2D pose estimation accuracy across all camera inputs for both estimators. For RTMO, the mAP drop when moving from normal to low lighting ranges from \textasciitilde 2.5--4.5\%, depending on the camera stream. For YOLO26, the reduction is generally smaller when applied to the three RGBD camera streams, at \textasciitilde 1--2\%, suggesting that YOLO26 is less affected by reduced illumination for these inputs.} 
% \mmn{again it seems very odd to say cameras are more robust to illumination for one method but not another method. The emphasis should be on the method and how it fares given a camera and its light-sensing sensitivity or output resolution etc.} 
{The main exception is the Logitech RGB stream, for which YOLO26 shows a larger decrease of \textasciitilde 5.4\%, indicating that the effect of lighting depends on both the estimator architecture and camera-specific RGB output characteristics, such as light sensitivity, resolution, and image quality. The introduction of furniture further reduces performance, with mAP drops of 3.0--4.5\% for RTMO and 3.0--7.7\% for YOLO26. For YOLO26, the largest occlusion-related reductions are observed on the ZED2 and Femto Bolt streams, suggesting greater sensitivity of this estimator to occlusion for these particular camera outputs.}
% ALREADY STATED ABOVE: , and the combined effect of low lighting and furniture produces the lowest scores across all cameras.

Camera mounting height has only a minor influence on 2D pose estimation within the evaluated range for both estimators. As shown in the rightmost panel of Fig.~\ref{fig:2D_pose}, mAP values remain relatively stable across the three mounting heights for all camera inputs under both lighting conditions, with variations typically below \textasciitilde 3.5 \%. For YOLO26, the variation remains below \textasciitilde 2.5 \% for every camera stream. This suggests that, within the tested height range, mounting height is not a dominant factor affecting 2D pose estimation accuracy. % compared with lighting, occlusion, and camera-specific RGB output characteristics.}

\vspace*{3mm}

\begin{figure*}[t]
    \centering
    \includegraphics[width=0.99\linewidth]{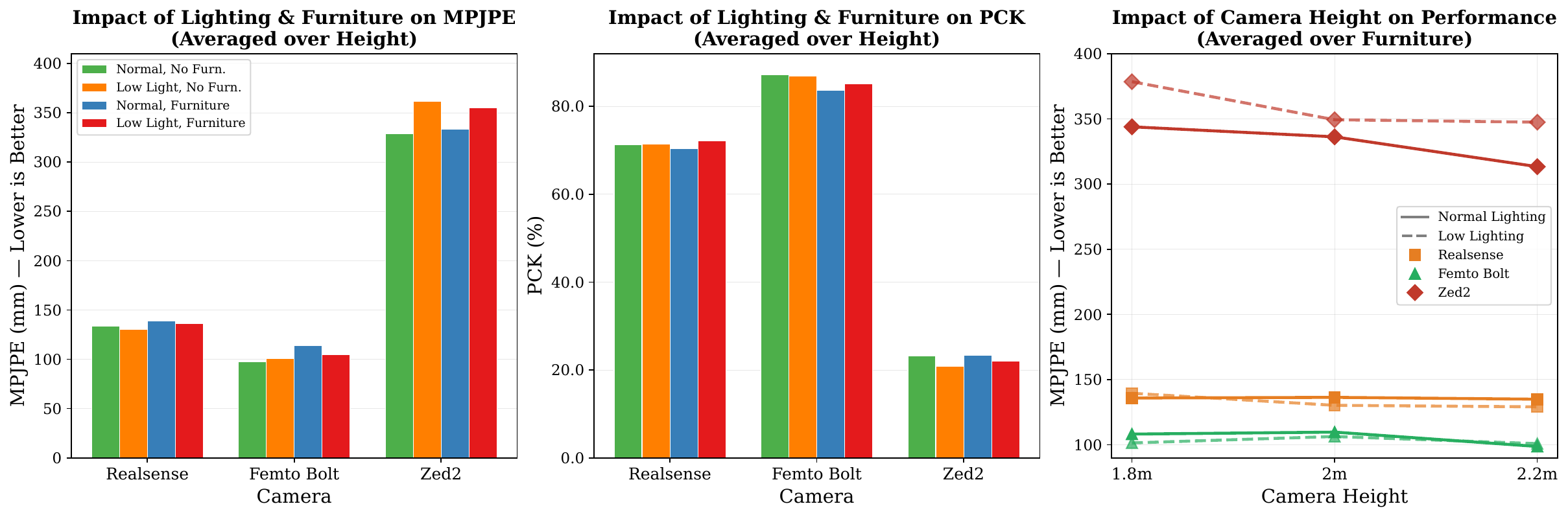}\\[2mm]
    {\small(\textbf{a}) RTMO}\\[3mm]
    \includegraphics[width=0.99\linewidth]{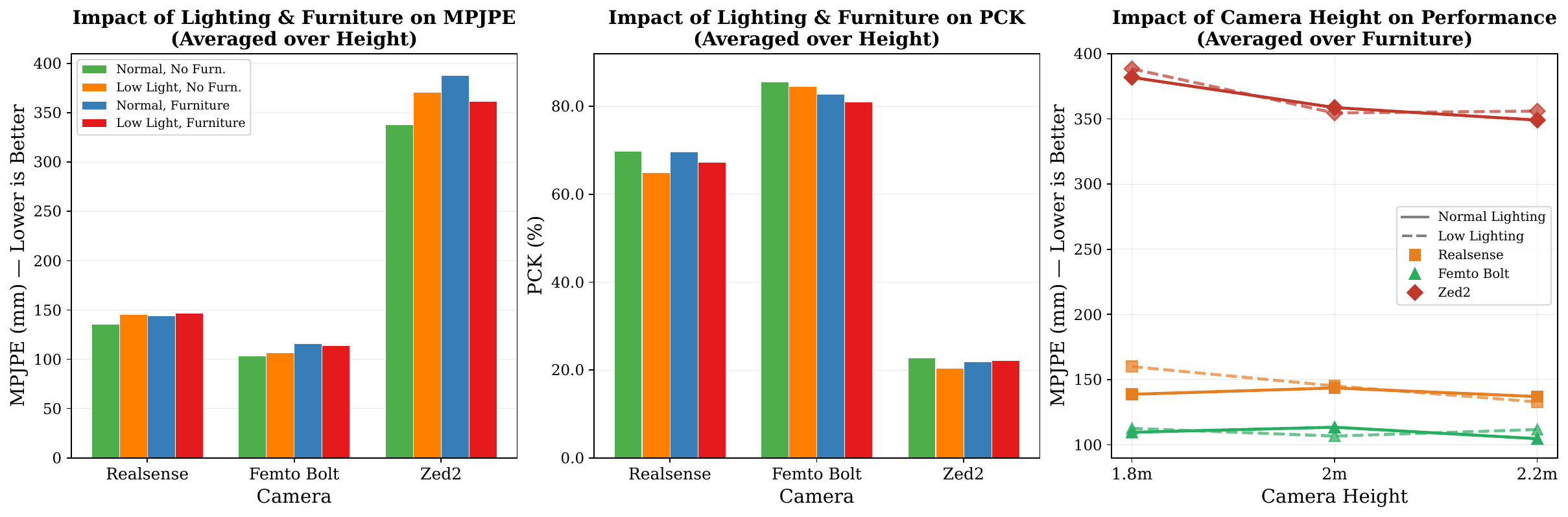}\\[2mm]
    {\small(\textbf{b}) YOLO26}
    \caption{{3D pose estimation performance across the three RGBD cameras for two pose
    estimators: (\textbf{a}) RTMO and (\textbf{b}) YOLO26. In each row, the left and
    centre panels show the impact of lighting and furniture occlusion on MPJPE and PCK
    respectively (averaged over mounting height), and the right panel shows the impact
    of camera mounting height on MPJPE (averaged over furniture). Lower MPJPE is better.}}
    \label{fig:3D_pose}
\end{figure*}

\noindent {\bf 3D pose estimation performance --}
Fig.~\ref{fig:3D_pose} shows the corresponding 3D pose estimation results for the
three RGBD camera streams. The 3D pose accuracy varies substantially across camera
inputs, reflecting differences in the underlying depth streams, with the same
device ordering and comparable error magnitudes observed for both RTMO and YOLO26.
{With RTMO, the lowest error is obtained for the Femto Bolt stream, with a mean
MPJPE of $104mm$ and a mean PCK of 85.7\%, indicating consistently reliable 3D pose
estimation across the evaluated conditions. The RealSense stream shows moderate
performance, with a mean MPJPE of $134mm$ and a mean PCK of 71.3\%, while the ZED2 stream
has substantially higher error, with a mean MPJPE of $345mm$ and a mean PCK of 22.3\%.
YOLO26 yields similar figures and the same ordering: Femto Bolt at a mean MPJPE of
$110mm$ and mean PCK of 83.5\%, RealSense at $143mm$ and 67.9\%, and ZED2 at $365mm$ and 21.8\%.
The close correspondence between the two architecturally distinct estimators supports
the interpretation that these differences are mainly associated with the
camera-specific depth streams rather than estimator-specific behaviour.} 
%\mmn{ideally, all th percentages should have at least 1 decimal point?}

The impact of environmental conditions is more pronounced for 3D than for 2D
evaluation, although for Femto Bolt and RealSense the absolute changes remain small
relative to their baseline error, and the largest effects are observed on the
ZED2 stream. Under normal lighting without furniture, all three cameras generally
achieve their best 3D performance. {The two estimators agree on the device ordering,
but differ in how the ZED2 stream responds to environmental change, so we report the
lighting and occlusion effects for each estimator in turn.} The transition to low
lighting causes a notable increase in MPJPE for the ZED2 stream {under RTMO, by
\textasciitilde $27mm$ on average over height and furniture}, while Femto
Bolt and RealSense remain largely unaffected, with average MPJPE changes below
$3mm$. {Under YOLO26, Femto Bolt again remains essentially unaffected (an average
change of about $1mm$) and RealSense shows a modest low-light increase of \textasciitilde
$6mm$. For the ZED2 stream, however, the low-light effect is strongly furniture
dependent: MPJPE increases by \textasciitilde 33~mm without furnitur,e but decreases by
\textasciitilde $26mm$ with furniture, so that the two nearly cancel to an average change
of only about $3mm$. This near-cancellation reflects the instability of ZED2's
depth-based 3D estimates under reduced illumination rather than genuine robustness to
lighting.} 

Furniture occlusion increases MPJPE on the Femto Bolt stream by \textasciitilde $10mm$ on
average and on the RealSense stream by \textasciitilde $5mm$, with both effects consistent
across the two estimators. {For the ZED2 stream, the occlusion effect is again
estimator dependent -- it adds little beyond the already high baseline error under RTMO,
whereas under YOLO26 it increases MPJPE by \textasciitilde $20mm$ on average.}

Camera height has limited impact on Femto Bolt and RealSense {under both
3D estimators}, with MPJPE values remaining relatively stable across the
$1.8$--$2.2$~m range {(for YOLO26, Femto Bolt varies by less than \textasciitilde
$3mm$)}. ZED2 shows gradual improvement at higher mounting positions (from $361mm$ at
$1.8$~m to $330mm$ at $2.2m$ {under RTMO, and from \textasciitilde $385mm$ at
$1.8m$ to \textasciitilde $353mm$ at $2.2m$ under YOLO26}), though the overall pose
estimation error remains considerably higher than that of the other two cameras.

\vspace*{3mm}

\noindent {\bf Summary --}
The 2D evaluation demonstrates that all four cameras provide reliable 2D pose
estimation under typical indoor conditions, with performance differences across
devices being relatively small {and the camera ranking preserved across both
estimators}. The 3D evaluation, however, reveals substantial differences driven
primarily by depth sensing quality, again with identical device ordering under
RTMO and YOLO26. The correspondence between metrological performance (Section \ref{Sec:metrological})
and 3D pose estimation accuracy is evident {under both estimators}: Femto Bolt,
which exhibited the lowest depth bias and highest precision, also achieves the best 3D
reconstruction, while ZED2's larger depth errors translate directly into higher MPJPE
values.
% \jj{This is kind of definite, we only use two pose estimators.}
These findings underline that for applications requiring 3D pose information, the
quality of the depth sensing modality is a critical factor. % than the 2D pose estimation algorithm.
Overall, Femto Bolt and RealSense demonstrate strong pose estimation performance, with
Femto Bolt achieving very low error. However, RealSense provides practical advantages
for system deployment, including a smaller form factor, wider field of view, and
support for 30fps depth acquisition compared to 15fps for Femto Bolt. RealSense also
offers broader software support and native compatibility with common computer vision
frameworks. We believe that the reported results can serve as a useful guide for those
seeking the best device to suit their own application.
\section{Discussion and Conclusion}
In this study, we proposed {two technical validation protocols} to evaluate RGB and RGBD camera performance under diverse metrological and indoor environmental conditions, and to assess their suitability for human movement monitoring indoors using pose estimation as a primary representation. {The protocols systematically evaluate five cameras (four RGBD and one RGB)} in the metrological assessment and four cameras (three RGBD and one RGB) in the pose estimation evaluation, across complementary dimensions: metrological characterisation of depth accuracy, thermal stability, temporal drift, and field of view, and then application-level assessment of 2D and 3D human pose estimation under controlled variations in lighting, camera placement, and occlusion.

The results demonstrate that camera performance is strongly influenced by environmental deployment factors. Reduced illumination and furniture occlusion both degrade pose estimation performance, though their relative impact varies across cameras and evaluation dimensions. In the 2D evaluation, lighting and furniture effects are of comparable magnitude across all cameras. In the 3D evaluation, the effect of lighting is camera-dependent: ZED2 shows substantial sensitivity to reduced illumination, whereas Femto Bolt and RealSense remain largely robust to lighting changes. The effect of furniture occlusion on 3D pose estimation also varies by camera and lighting condition, with no single pattern applying uniformly across all devices. Camera mounting height, within the $1.8-2.2m$ range, has a comparatively minor effect on pose estimation, suggesting that height selection can be guided primarily by field-of-view coverage rather than accuracy considerations. The strong correspondence between metrological and pose estimation results highlights the value of the two-stage approach, enabling application-level degradation to be traced back to sensor-level characteristics.

Rather than identifying a single universally optimal device, the findings highlight that camera suitability depends on deployment requirements. Cameras differ in their trade-offs between spatial coverage, measurement stability, and pose reconstruction reliability. For applications requiring accurate 3D pose reconstruction, Femto Bolt offers the best performance, while RealSense provides a balanced compromise between accuracy and deployment practicality. For 2D-only applications, all four cameras, including the RGB-only Logitech, provide comparable detection performance, allowing selection to be driven by cost and integration requirements. 
{In terms of cost, the RGB-only Logitech BRIO~4K is naturally lower-priced than RGBD devices which occupy a relatively comparable price band. As  absolute prices vary by vendor, region and changes over time, we will not report them.}

%\ame{In terms of relative cost, the RGB-only Logitech BRIO~4K occupies the lowest price tier, whereas the RGBD devices are in a higher and broadly comparable price band. As  absolute prices vary by vendor, region and changes over time, we report only this relative positioning.\footnote{Prices are indicative at the time of writing and are intended only for relative comparison across device classes rather than as procurement figures; current vendor quotations should be consulted before purchase.} This cost difference is most relevant for 2D-only deployments, where the comparable 2D pose estimation accuracy of all four cameras means that the substantially cheaper RGB webcam may be sufficient, whereas applications that require reliable 3D pose information must accept the higher cost of an RGBD sensor in exchange for depth sensing.}

{Several limitations should be acknowledged. The pose estimation experiments revolved around a single participant, ensuring consistency, but limiting assessment of inter-subject variability. Two pose estimation methods, RTMO~\cite{lu2024rtmo} and YOLO26~\cite{sapkota2025yolo26}, were used, selected as state-of-the-art single-stage real-time estimators suited to edge deployment; evaluating two architecturally distinct methods allows the camera-level conclusions to be checked for robustness against estimator-specific behaviour. Other estimation paradigms were not included in the full evaluation: several alternatives were assessed in preliminary experiments, including MediaPipe~\cite{lugaresi2019mediapipe} and AlphaPose~\cite{fang2022alphapose}, but were not retained, and heavier offline or top-down estimators fell outside the real-time scope of this study. Consequently, relative camera performance may still differ under other algorithms. The occlusion conditions were limited to a fixed furniture arrangement, whereas real-world environments naturally exhibit greater variability. Future work should extend the evaluation to multiple participants, incorporate additional pose estimation methods, and assess performance under more varied occlusion scenarios.}
%Overall, the proposed protocol provides a structured and reproducible framework for evaluating camera reliability and supports informed selection of sensing systems for real-world vision-based mobility monitoring and digital biomarker analysis.
%\section*{Acknowledgments}
%This work was supported by the TORUS Project, which has been funded by the UK Engineering and Physical Sciences Research Council (EPSRC), grant number EP/X036146/1. The study was conducted at the Bristol Robotics Laboratory (BRL). Ethical approval was obtained under the University of Bristol Ethics framework. 

%%%%%%%%%%%%%%%%%%%%%%%%%%%%%%%%%%%%%%%%%%
\vspace{6pt} 

%%%%%%%%%%%%%%%%%%%%%%%%%%%%%%%%%%%%%%%%%%
%% optional
%\supplementary{The following supporting information can be downloaded at:  \linksupplementary{s1}, Figure S1: title; Table S1: title; Video S1: title.}


% Only used for preprtints:
% \supplementary{The following supporting information can be downloaded at the website of this paper posted on \href{https://www.preprints.org/}{Preprints.org}.}

%\begin{tabularx}{\textwidth}{lll}
%\toprule
%\textbf{Name} & \textbf{Type} & \textbf{Description} \\
%\midrule
%S1 & Python script (.py) & Script of python source code used in XX \\
%S2 & Text (.txt) & Script of modelling code used to make Figure X \\
%S3 & Text (.txt) & Raw data from experiment X \\
%S4 & Video (.mp4) & Video demonstrating the hardware in use \\
%\bottomrule
%\end{tabularx}
%}

%%%%%%%%%%%%%%%%%%%%%%%%%%%%%%%%%%%%%%%%%%
\authorcontributions{Conceptualization,  Majid Mirmehdi; methodology, Amirhossein Dadashzadeh, Jingjing Liu, and Majid Mirmehdi; software, Amirhossein Dadashzadeh and Jingjing Liu; validation, Amirhossein Dadashzadeh and Jingjing Liu; formal analysis,  Jingjing Liu and Amirhossein Dadashzadeh; investigation, Qianhui Men and Qiushuo Cheng; resources, Majid Mirmehdi, Kirsty Scott, and Lisa Alcock; data curation, Amirhossein Dadashzadeh and Jingjing Liu; writing---original draft preparation,  Amirhossein Dadashzadeh and Jingjing Liu; writing---review and editing, Majid Mirmehdi; visualization, Qianhui Men, Jingjing Liu, Amirhossein Dadashzadeh; supervision, Majid Mirmehdi and Ian Craddock; project administration, Ian Craddock; funding acquisition, Ian Craddock (PI) and Majid Mirmehdi (Co-I). All authors have read and agreed to the published version of the manuscript.} %please turn to the  \href{http://img.mdpi.org/data/contributor-role-instruction.pdf}{CRediT taxonomy} for the term explanation. Authorship must be limited to those who have contributed substantially to the work~reported.

\funding{This work was supported by the TORUS Project, which has been funded by the UK Engineering and Physical Sciences Research Council (EPSRC), grant number EP/X036146/1.} %The research was also supported by the National Institute for Health and Care Research (NIHR) Newcastle Biomedical Research Centre based at The Newcastle upon Tyne Hospitals NHS Foundation Trust, Newcastle University and the Cumbria, Northumberland and Tyne and Wear (CNTW) NHS Foundation Trust.

\institutionalreview{ Ethical approval was obtained under the University of Bristol Ethics framework.}

\informedconsent{Informed consent was obtained from all participants involved in the study.

%Any research article describing a study involving humans should contain this statement. Please add ``Informed consent was obtained from all subjects involved in the study.'' OR ``Patient consent was waived due to REASON (please provide a detailed justification).'' OR ``Not applicable'' for studies not involving humans. You might also choose to exclude this statement if the study did not involve humans.
%Written informed consent for publication must be obtained from participating patients who can be identified (including by the patients themselves). Please state ``Written informed consent has been obtained from the patient(s) to publish this paper'' if applicable.
}

\dataavailability{
The datasets presented in this article are not publicly available due to time limitations. Access to the datasets may be considered on a reasonable request and subject to approval by the authors and the host institution. Requests should be directed to the corresponding author.}

\acknowledgments{The authors would like to thank the Bristol Robotics Laboratory (BRL) for providing the facilities used to conduct a substantial part of the experimental work. We are particularly grateful to Patrick Brinson and Jakub Jezierski for their technical support and assistance during the experiments.}

\conflictsofinterest{The authors declare no conflicts of interest.} 

%%%%%%%%%%%%%%%%%%%%%%%%%%%%%%%%%%%%%%%%%%
%% Optional

%% Only for journal Encyclopedia
%\entrylink{The Link to this entry published on the encyclopedia platform.}

\abbreviations{Abbreviations}{
The following abbreviations are used in this manuscript:
\\

\noindent 

\begin{tabular}{@{}ll}
RGB & Red Green Blue\\
RGBD & Red Green Blue and Depth\\
FOV & Field of View\\
HFOV & Horizontal Field of View\\
VFOV & Vertical Field of View\\
DFOV & Diagonal Field of View\\
ROI & Region of Interest\\
RMSE & Root Mean Square Error\\
AAD & Average Absolute Drift\\
IMU & Inertial Measurement Unit\\
OKS & Object Keypoint Similarity\\
mAP & Mean Average Precision\\
PCK & Percentage of Correct Keypoints\\
MPJPE & Mean Per-Joint Position Error\\
MoCap & Motion Capture\\
ICC & Intraclass Correlation Coefficient\\
CMC & Coefficient of Multiple Correlation\\
TUG & Timed Up and Go\\
FPS & Frames Per Second
\end{tabular}

}

\newpage
%%%%%%%%%%%%%%%%%%%%%%%%%%%%%%%%%%%%%%%%%%
%% Optional
\appendixtitles{no} % Leave argument "no" if all appendix headings stay EMPTY (then no dot is printed after "Appendix A"). If the appendix sections contain a heading then change the argument to "yes".
\appendixstart
\appendix
\section[\appendixname~\thesection]{HFOV and VFOV}
\label{appendix_fov}
\begin{figure}[!htp]
    \centering
    \begin{subfloat}[]{
        \includegraphics[width=0.95\columnwidth]{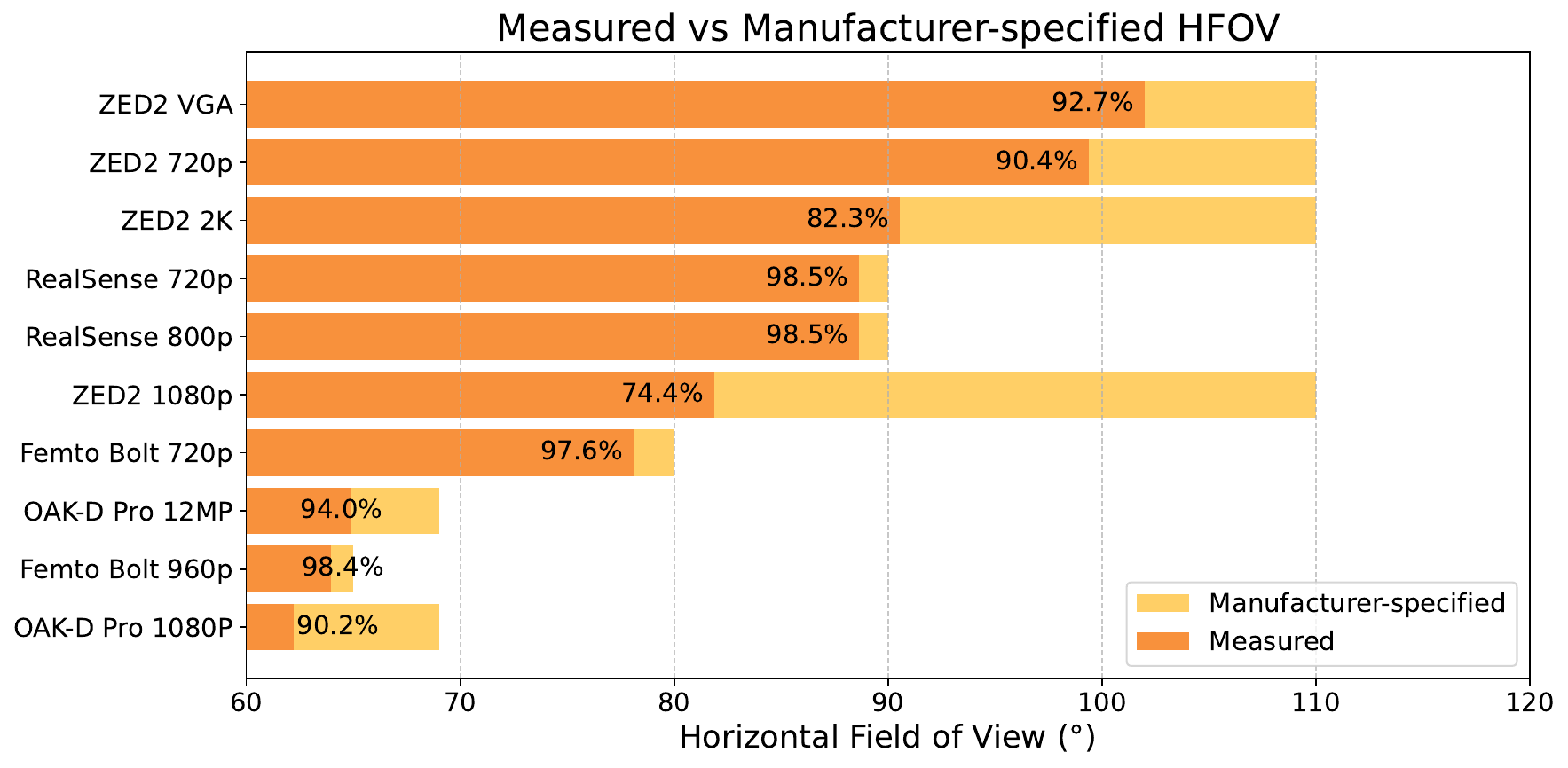}}
    \end{subfloat}
    \begin{subfloat}[]{
        \includegraphics[width=0.95\columnwidth]{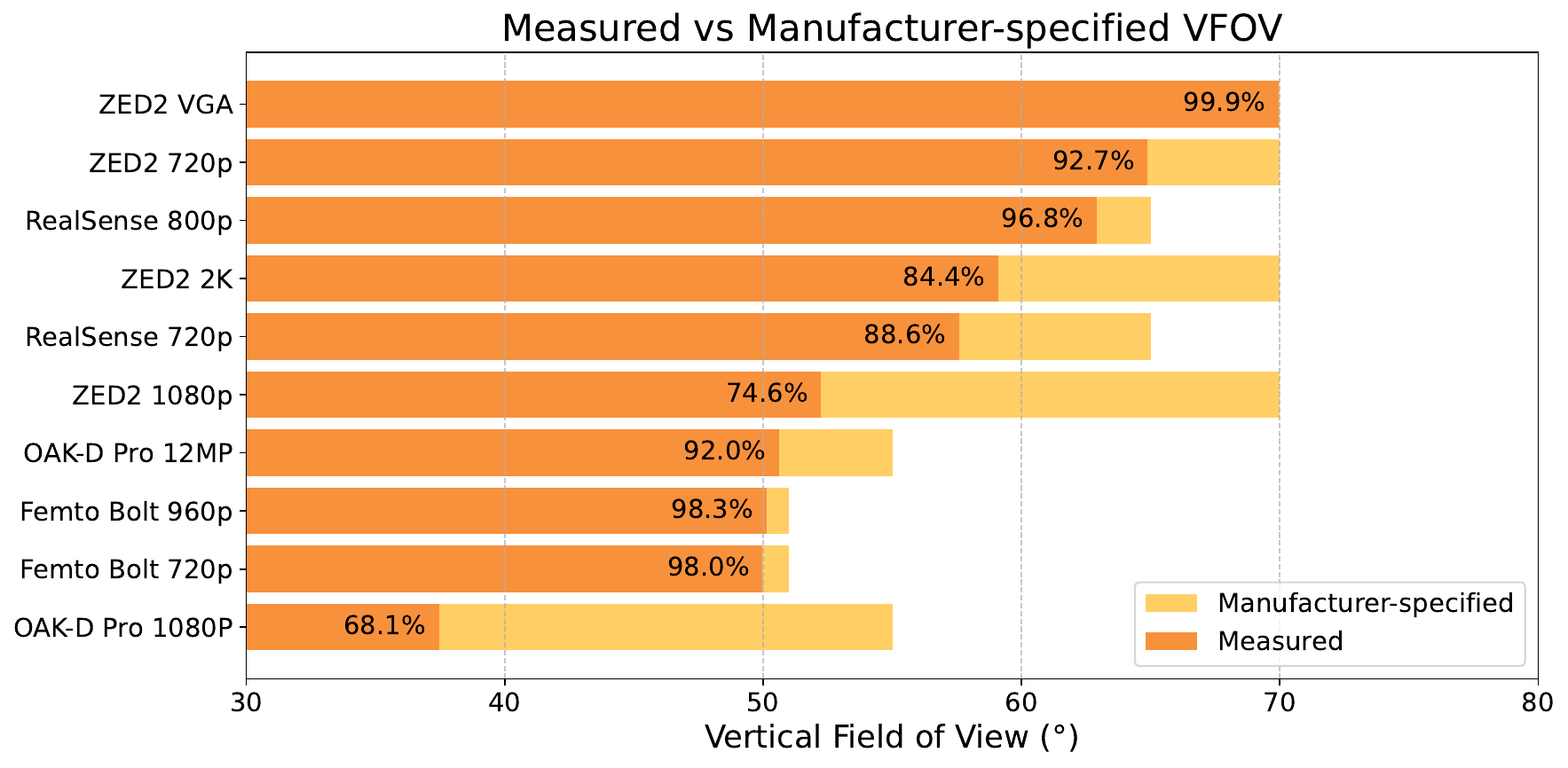}}
    \end{subfloat}
    \caption{Comparison of measured and manufacturer-specified (a) Horizontal Field of View (HFOV) and (b) Vertical Field of View (VFOV) across multiple camera models and resolutions. }
\label{fig:hv_fov}
\end{figure}

The results of HFOV and VFOV of all camera–resolution combinations against their manufacturer specifications are presented in Fig.~\ref{fig:hv_fov}. 
%The results of DFOV angles under different resolutions are compared in Fig.~\ref{fig:fov_cameras}.

% FOV, HOV

\section[\appendixname~\thesection]{Synchronisation}
Temporal synchronisation between the RGB/RGBD cameras and the Vicon motion capture system is achieved using a trigger-based event, specifically a clap action performed by the participant within the scene prior to the motion tasks. The clap event is detected by identifying the timestamp corresponding to the minimum distance between the left and right wrists. This timestamp is denoted as $t_{V,\mathrm{clap}}$ for the Vicon system and $t_{C,\mathrm{clap}}$ for the camera. Let $f_V$ and $f_C$ denote the frame rates of the Vicon system and the camera, respectively. The Vicon timestamps aligned to the camera time reference are then computed as
\begin{equation}
t_{V\to C}(i) = t_{C,\mathrm{clap}} + \frac{f_C}{f_V}\left(t_V(i) - t_{V,\mathrm{clap}}\right).
\end{equation}
Given the aligned timestamps $t_{V\to C}(i)$, one-dimensional interpolation is applied to resample the Vicon data, yielding temporally synchronized Vicon trajectories corresponding to the camera frames.

\section{Intrinsic and extrinsic calibration}
Intrinsic calibration was performed for all cameras using a standard chessboard-based calibration procedure \cite{zhang1999flexible}. For each camera, the extrinsic calibration between the camera coordinate system and the Vicon motion capture coordinate system was then conducted, allowing kinematic data from both systems to be expressed in a common global coordinate frame.

\begin{figure}[!htp]
    \centering
        \includegraphics[width=0.6\columnwidth]{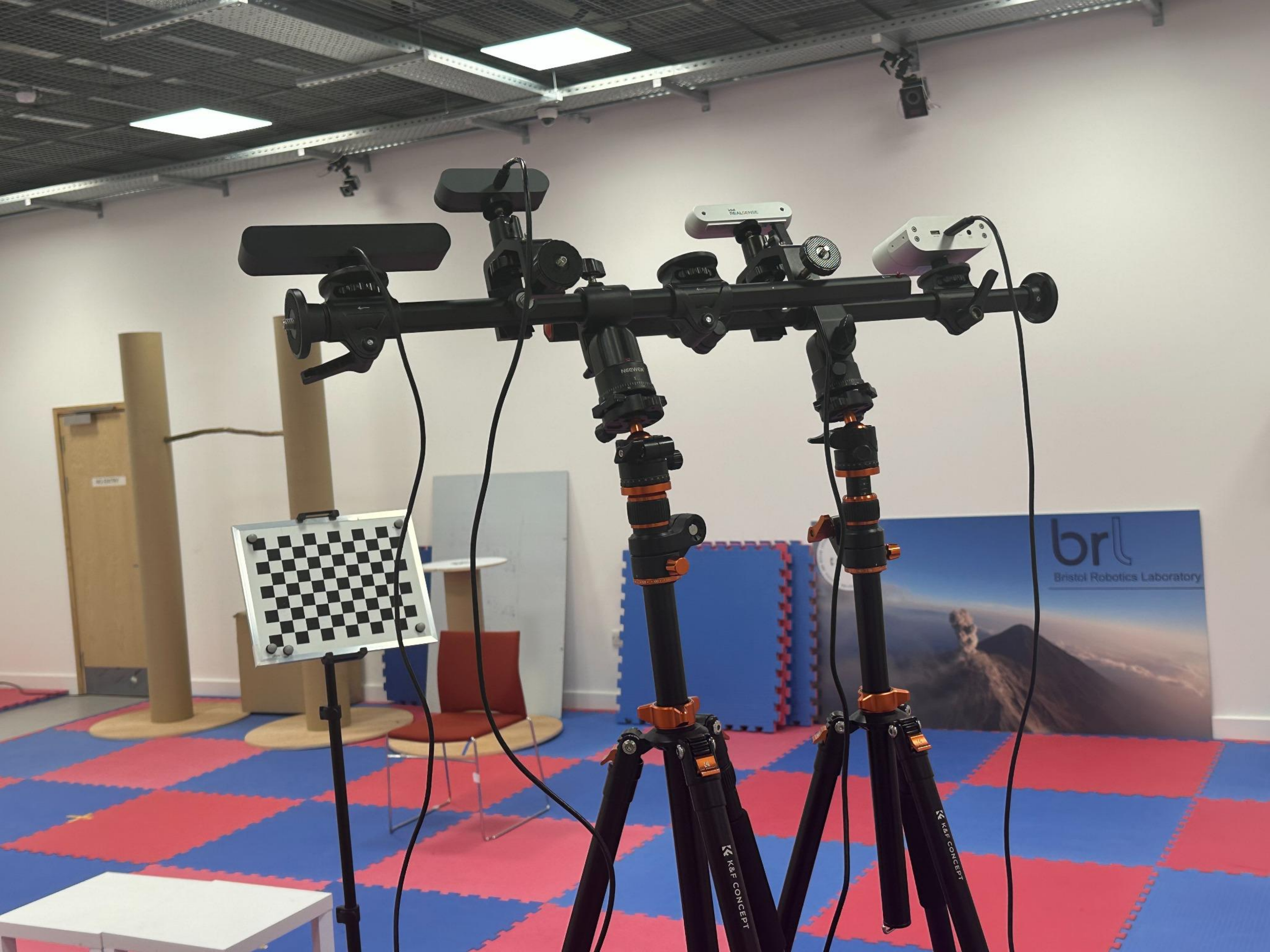}
    \caption{Calibration setup between MoCap and RGB/RGBD cameras.}
\label{fig:calibration}
\end{figure}

The extrinsic calibration setup employed a chessboard with four retro-reflective markers attached to its outer corners. The chessboard was positioned such that it was simultaneously visible to both the camera and the Vicon system, and synchronized recordings of camera images and marker positions were acquired (see Fig.~\ref{fig:calibration}). This procedure was repeated ten times with the chessboard placed at different positions and orientations to ensure sufficient calibration diversity. {For each time, we record the image of the chessboard using the camera and the 3D positions of markers on the chessboard using the Vicon system.} %\amn{if we are gonna mention about calibration we need to put some other details here as well. For example what did we do with these ten calibration? i think better we mention here that we just did calibration and details in appendix or supp. It's not a part of our camera evaluation.}

For each recorded image, the 2D locations of the chessboard inner corners were detected using OpenCV's \texttt{findChessboardCorners()} function. The corresponding 3D coordinates of these corners were inferred from the four outer corner markers captured by the Vicon system, using bilinear interpolation based on the known chessboard geometry. This resulted in a set of 2D–3D point correspondences between the camera image plane and the Vicon coordinate system. The camera extrinsic parameters were subsequently estimated using OpenCV's \texttt{solvePnP()} function, enabling the transformation of vicon-based skeleton data into the camera's coordinate system.

%%%%%%%%%%%%%%%%%%%%%%%%%%%%%%%%%%%%%%%%%%
%\isPreprints{}{% This command is only used for ``preprints''.
\begin{adjustwidth}{-\extralength}{0cm}

\reftitle{References}

% --- Tight reference list spacing ---
\makeatletter
\renewenvironment{thebibliography}[1]
{%
  \par\vspace{12pt}
  \noindent\hspace*{3cm}{\fontsize{12}{12}\selectfont\bfseries References}%
  \par\vspace{6pt}
  \fontsize{9}{11}\selectfont
  \list{\arabic{enumi}.}
  {%
    \usecounter{enumi}
    \setlength{\labelwidth}{0.35cm}
    \setlength{\labelsep}{0.15cm}
    \setlength{\leftmargin}{3.55cm}
    \setlength{\itemindent}{0pt}
    \setlength{\itemsep}{2pt}
    \setlength{\parsep}{0pt}
    \setlength{\topsep}{0pt}
    \renewcommand{\makelabel}[1]{\hss##1}
  }%
  \sloppy
}
{%
  \endlist
}
\makeatother

\bibliography{reference}
\end{adjustwidth}
\end{document}